\documentclass[10pt,twocolumn,letterpaper,table]{article}
\pdfoutput=1

\usepackage[table]{xcolor}
\usepackage[pagenumbers]{cvpr}              
\usepackage[accsupp]{axessibility}

\makeatletter
\@namedef{ver@everyshi.sty}{}
\makeatother

\usepackage{comment}
\usepackage{amsmath,amssymb} 
\usepackage{xurl} 
\usepackage{times}
\usepackage{epsfig}
\usepackage{graphicx}
\usepackage{color}
\usepackage{bm}
\usepackage{multirow}
\usepackage{diagbox}
\usepackage{array}
\usepackage{booktabs}
\usepackage{nth}
\usepackage{nicefrac}
\usepackage{relsize}
\usepackage{xspace} 
\usepackage{bm}
\usepackage{enumerate}
\usepackage{adjustbox}
\usepackage{enumitem}
\usepackage{array}
\usepackage{cuted}
\usepackage{capt-of}
\usepackage[accsupp]{axessibility}
\usepackage{booktabs}
\usepackage{overpic}
\usepackage{cellspace, tabularx}
\usepackage{dcolumn}
\usepackage{xr}
\newcolumntype{G}{D{,}{,}{2.1}}

\definecolor{superlightgray}{gray}{0.92}
\definecolor{lightgray}{gray}{0.8}
\definecolor{mediumgray}{RGB}{160,170,180}
\definecolor{azul}{RGB}{230,240,250}
\newcommand{\cc}{\cellcolor{azul}}
\newcommand{\gr}{\cellcolor{superlightgray}}

\usepackage{tcolorbox}
\newtcbox{\inlinecode}{on line, boxrule=0pt, boxsep=0pt, top=1pt, left=2pt, bottom=1pt, right=2pt, colback=lightgray!15, colframe=gray, fontupper={\ttfamily \footnotesize}}

\usepackage[pagebackref,breaklinks,colorlinks]{hyperref}

\usepackage[capitalize]{cleveref}
\crefname{section}{Sec.}{Secs.}
\Crefname{section}{Section}{Sections}
\Crefname{table}{Table}{Tables}
\crefname{table}{Tab.}{Tabs.}

\newcommand\blfootnote[1]{%
  \begingroup
  \renewcommand\thefootnote{}\footnote{#1}%
  \addtocounter{footnote}{-1}%
  \endgroup
}

\def\cvprPaperID{6370}
\def\confName{CVPR}
\def\confYear{2023}

\newcommand{\name}{VIVE3D\xspace}
\newcommand{\SIT}{StiiT\xspace}
\newcommand{\VEG}{VEG\xspace}
\renewcommand{\vec}[1]{\ensuremath{\mathbf{#1}}}
\newcommand{\generator}[1]{\ensuremath{\mathcal{G}_{#1}}} 
\newcommand{\gtuned}{\ensuremath{\generator{\mtxt{ID}}}\xspace} 
\newcommand{\mtxt}[1]{\text{\textit{#1}}} 
\newcommand{\w}{\vec{w}\xspace} 
\newcommand{\wperson}{\ensuremath{\vec{w}_\mtxt{ID}}\xspace} 
\newcommand{\idxS}{\mtxt{n}\xspace} 
\newcommand{\idxF}{\mtxt{f}\xspace} 
\newcommand{\source}{\ensuremath{S}\xspace} 
\newcommand{\face}[1]{\ensuremath{\vec{F}_{#1}}\xspace} 
\newcommand{\seg}[2]{\ensuremath{\vec{S}_{\mtxt{#1}}({#2})}\xspace} 
\newcommand{\down}[2]{\ensuremath{\vec{D}_{\mtxt{#1}}({#2})}\xspace} 
\newcommand{\offset}[1]{\ensuremath{\vec{o}_{#1}}\xspace} 
\newcommand{\loss}[1]{\ensuremath{\mathcal{L}_{\mtxt{#1}}}\xspace}
\newcommand{\weight}[1]{\ensuremath{\lambda_{\mtxt{#1}}}\xspace}
\newcommand{\y}[1]{\ensuremath{\mtxt{yaw}_{#1}}\xspace} 
\newcommand{\cam}[1]{\ensuremath{c_{#1}}\xspace} 
\newcommand{\p}[1]{\ensuremath{\mtxt{pitch}_{#1}}\xspace} 
\newcommand{\edit}{\mtxt{edit}\xspace} 

\newcommand{\figlabel}[1]{\textbf{(#1)}}

\newcommand\mr[2]{\multirow{#1}{*}{#2}}
\newcommand\mc[2]{\multicolumn{#1}{c}{#2}}
\newcommand\tss[1]{\textsc{#1}}
\newcommand\mins{{\scriptsize m~}}
\newcommand\secs{{\scriptsize s}}
\newcommand\pgraph[1]{\noindent\textbf{{#1}.}}

\newcommand{\mytilde}{\raise.17ex\hbox{$\scriptstyle\mathtt{\sim}$}}

\begin{document}

\title{\name: Viewpoint-Independent Video Editing using 3D-Aware GANs}
\author{
    Anna Fr\"{u}hst\"{u}ck\textsuperscript{{2*}}, 
    Nikolaos Sarafianos\textsuperscript{1},
    Yuanlu Xu\textsuperscript{1}, 
    Peter Wonka\textsuperscript{2}, 
    Tony Tung\textsuperscript{1}
\\[0.4ex]
	\textsuperscript{1~}Meta Reality Labs Research, Sausalito \quad
        \textsuperscript{2~}KAUST \quad \\ 
        \small{\tt\href{https://afruehstueck.github.io/vive3D}{afruehstueck.github.io/vive3D}}
}


\setlength{\abovedisplayskip}{4pt}
\setlength{\belowdisplayskip}{4pt}
\setlength{\belowcaptionskip}{-5pt}

\twocolumn[{%
\renewcommand\twocolumn[1][]{#1}%
\vspace*{-.5in}
\maketitle
\vspace*{-.4in}
\begin{center}
  \centering
  \captionsetup{type=figure}
  \includegraphics[width=0.85\textwidth]{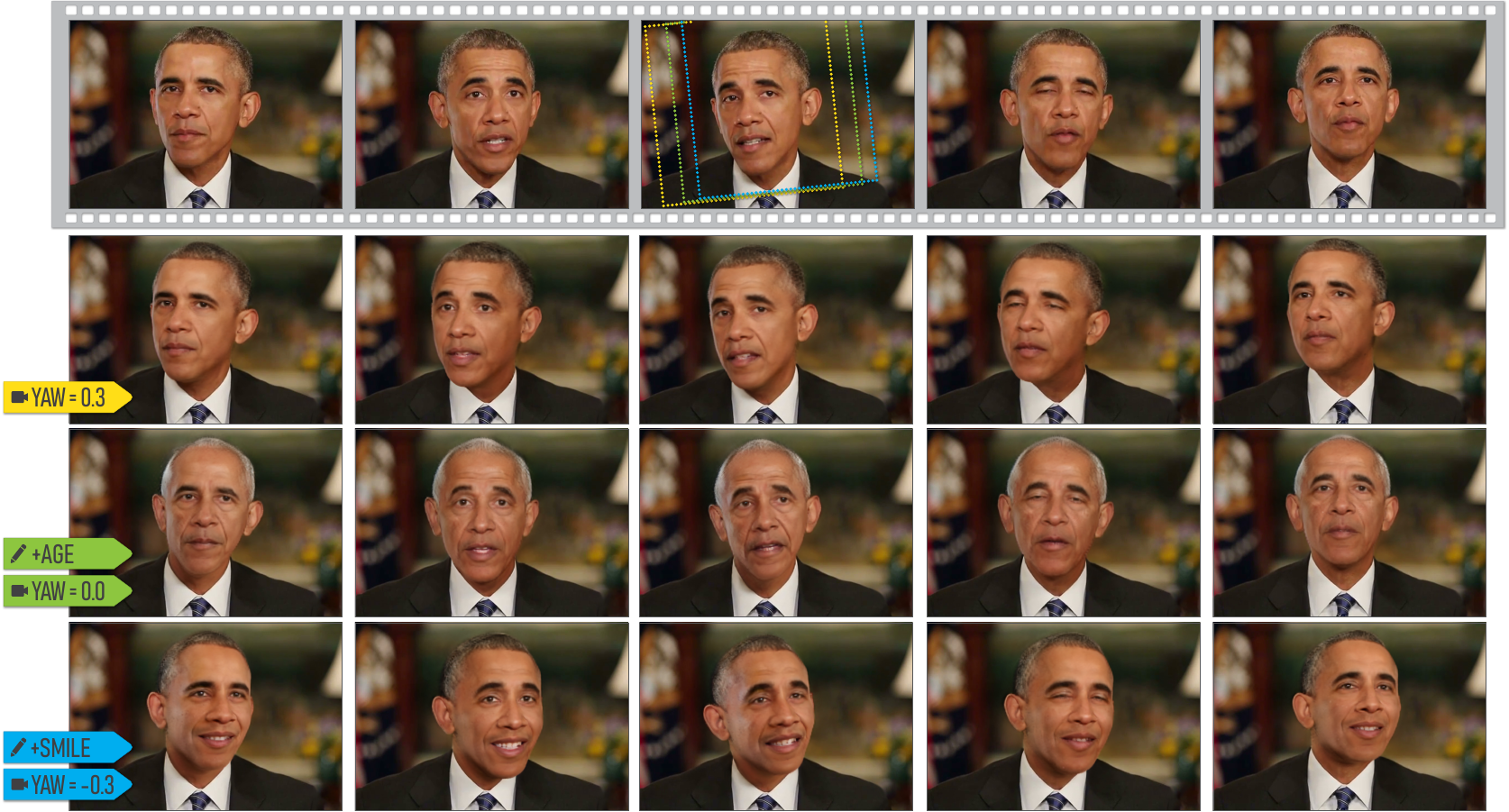}
  \vspace*{-.07in}
  \caption{We propose \textbf{\name}, a novel method that creates a powerful personalized 3D-aware generator using a low number of selected images of a target person. 
  Given a new video of that person, we can faithfully modify several facial attributes as well as the camera viewpoint of the head crop. Finally, we seamlessly composite the edited face with the source frame in a temporally and spatially consistent manner, while retaining a plausible composition with the static components of the frame outside of the generator's region. The dotted squares in the center frame denote the reference regions for the three different camera poses in the column below.}
  \label{fig:teaser}
\end{center}%
}]

\begin{abstract}
We introduce \name, a novel approach that extends the capabilities of image-based 3D GANs to video editing and is able to represent the input video in an identity-preserving and temporally consistent way.
We propose two new building blocks.
First, we introduce a novel GAN inversion technique specifically tailored to 3D GANs by jointly embedding multiple frames and optimizing for the camera parameters.
Second, besides traditional semantic face edits (\eg for age and expression), we are the first to demonstrate edits that show novel views of the head enabled by the inherent properties of 3D GANs and our optical flow-guided compositing technique to combine the head with the background video.
Our experiments demonstrate that \name generates high-fidelity face edits at consistent quality from a range of camera viewpoints which are composited with the original video in a temporally and spatially consistent manner. 

\blfootnote{*This work was conducted during an internship at Meta RL Research.}
\end{abstract}


\section{Introduction}
\vspace{-1mm}
\label{sec:intro}
Semantic image editing has been an active research topic for the past few years. Previous work~\cite{Goodfellow2014GAN} uses Generative Adversarial Networks (GANs) to produce high-fidelity results in the image space. The most popular backbone is StyleGAN~\cite{Karras2020StyleGAN2, Karras2019StyleGAN, Karras2020StyleGAN2ADA, Karras2021StyleGAN3} as it generates high-resolution domain-specific images while providing a disentangled latent space that can be utilized for editing operations.
To edit real photographs, there are typically two steps: The first step maps the input image to the latent space of a pre-trained generator. This is usually accomplished either through encoder-based embedding or through optimization, such that generator can accurately reconstruct the image from the latent code~\cite{xia2022gan}. The second step is semantic image manipulation, where one latent input representation is mapped to another to obtain a certain attribute edit, (\eg changing age, facial expression, glasses, or hairstyle).
While existing approaches produce impressive results on single images, extending them to videos is far from straightforward. Among the challenges that arise are: (1) people tend to move their heads freely in videos (instead of assuming frontal image inputs), (2) the inversion of multiple frames should be coordinated, (3) the inverted face and edits need to be temporally consistent and (4) the compositing of the edited face with the original frame must maintain boundary consistency. 

A recent set of approaches has focused on 3D-aware GANs where a 2D face generator is combined with a neural renderer. Given a latent code, a 2D image and the underlying 3D geometry are generated, thus allowing for some camera movement while rendering the head of the person.

In this paper, we tackle the problem of viewpoint-independent face editing in videos. The edited face is rendered from novel views in a temporally-consistent manner. Specifically, we use a 3D-aware GAN in the temporal domain and apply facial image editing techniques per frame that are temporally smooth regardless of the rendered view.
Compared with other GAN-based video editing approaches~\cite{Tzaban2022STIT, Alaluf2022Times}, our method is the first to perform viewpoint-independent video editing while showing the full upper body of the person in the video with high fidelity.

\name takes a video of a person captured from a monocular camera as input. The captured person can move freely across time, talk, and make facial expressions while their body can be visible. Unlike all prior work that learns a generator and performs edits on the exact same video, we disentangle these steps. Hence the output of our approach can be a different video of the same person or the same video. In both cases, the face has undergone one or more attribute edits and is rendered from a novel view. To accomplish this challenging task, we introduce several novel components, each addressing one challenge of the problem at hand. Specifically, we first propose a simple yet effective technique to create a personalized generator by inverting multiple frames at the same time. The simultaneous inversion of \(N\) frames exposes the generator to a variety of facial poses and expressions, which results in a larger capacity that we can then utilize. Our generator can generalize to new unseen videos of the same identity where the person might be wearing a different shirt, a result not demonstrated in the literature so far. In addition, we propose to optimize the camera pose of the 3D-aware GAN during inversion to obtain an accurate estimate which angle the face was captured from. Finally, we introduce an optical flow-based compositing method to properly place the novel view of the edited face back into the original frame while ensuring that the end result is temporally and spatially consistent. 
Our experimental work provides a wide range of qualitative and quantitative results to demonstrate that \name accomplishes semantic video editing with changing camera poses in a faithful way. In summary, our contributions are: 

\begin{itemize}[leftmargin=*]
\itemsep-0.1em 
    \item A new 3D GAN inversion technique that jointly embeds multiple images while optimizing for their camera poses.
    \item A complete attribute editing framework and an optical flow-based compositing technique to replace the edited face in the original video.
    \item \name is the first 3D GAN-based video editing method and the first that can change the camera pose of the face.
\end{itemize}

\begin{figure*}[t]
\centering
\includegraphics[width=\linewidth]{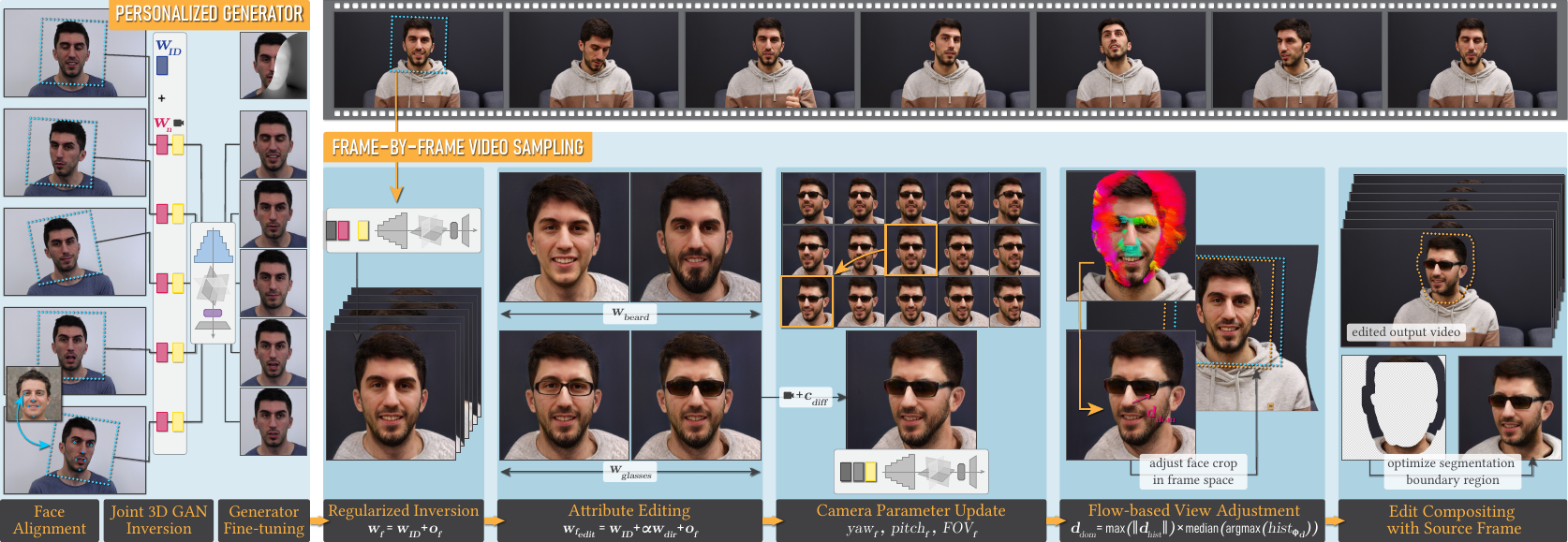}
   \vspace{-6mm}
   \caption{\textbf{\name Pipeline.} 
    To create an edited video, we first need to create a personalized generator by jointly inverting selected faces and fine-tuning a pre-trained generator. We then invert the cropped face regions from a source video (which could be the same or a different video) into our personalized generator and recover the latent codes and camera poses for each target frame. We are able to perform semantic editing on the inverted stack of latent codes using previously discovered latent space directions and we can freely change the camera path around the face region. In order to composite the face with the source frame in a consistent fashion, we use optical flow to correct the position of the inset within the frame, which allows us to composite the result in a seamless and temporally consistent fashion. 
    }
   \label{fig:pipeline}
   \vspace{-4mm}
\end{figure*}

\vspace{-0.45cm}
\section{Related Work}\label{sec:related_work}
\vspace{-0.1cm}
\pgraph{GAN Inversion}
GANs are a powerful tool for semantic editing. Most editing techniques are tailored to StyleGAN, the state-of-the-art of {2D} GANs~\cite{Karras2019StyleGAN, Karras2020StyleGAN2, Karras2020StyleGAN2ADA, Karras2021StyleGAN3}. Several editing techniques~\cite{Fruehstueck2022InsetGAN, Fruehstueck2019TileGAN,karakas2022fairstyle, chaudhuri2021semi, parmar2022spatially} build upon StyleGAN as it uses an intermediate disentangled latent space, usually referred to as \w-space.
Before editing, a latent space representation of the input image has to be recovered using a process typically referred to as Inversion or Projection~\cite{Creswell2018Inverting, Abdal2019Image2StyleGAN, Abdal2020Image2StyleGAN++, Wang2022CVPR}. Refer to~\cite{Xia2022InversionSurvey} for a survey of inversion techniques. In contrast to optimization-based inversion techniques, learning-based approaches attempt to obtain faster latent space correspondences by training encoders~\cite{richardson2021encoding, tov2021designing, alaluf2021restyle}. 
In order to retain the generalization ability of the \w-space while providing a high-quality inversion, Pivotal Tuning~\cite{Roich2022PTI} has successfully shown that trained generators can overfit to target images while still maintaining a navigable latent space.
Recent works study 3D GAN inversion~\cite{Lin20223DGANInversion, Ko20233D}, attempting to infer a 3D representation for a reference image.

\pgraph{GAN-based Latent Space Editing}
Once an appropriate latent space representation of an input image has been recovered, semantic edits can be applied by navigating the latent space manifold surrounding the inverted latent code. Unsupervised techniques attempt to find interesting edits without labeled data~\cite{Jahanian2019Steerability, Harkonen2020GANSpace, Shen2021SeFa, Voynov2020Unsupervised}.
InterfaceGAN~\cite{Shen2020Interpreting, Shen2020InterFaceGAN} is a simple and robust supervised technique that is highly recommended for practical applications, and as such we also employ it in our work.
While there is a plethora of other techniques~\cite{Zhuang2021ICLREditing, Yao2021LatentTransformer, Cherepkov2021Navigating, Abdal2021StyleFlow, Tewari2020StyleRig, Tewari2020Pie} the development of related latent space manipulations itself is not the focus of our work. Another line of work is text-based editing which gained immense popularity during the last year~\cite{Gal2022StyleGANNADA, Patashnik2021StyleClip}.

\pgraph{3D-aware GANs}
Recent GAN papers attempt to discover 3D information from large collections of 2D images using Neural Radiance Fields (NeRFs) as shape representations~\cite{Chan2021EG3D,Orel2022StyleSDF,Deng2022GRAM,Hong2022EVA3D,Zhao2022GMPI, Schwarz2022NEURIPS,Cai2022pix2nerf}. While most of these papers share similar architectural ideas, EG3D~\cite{Chan2021EG3D} has emerged as a popular basis for follow-up work (\eg integration of a segmentation branch~\cite{Sun2022IDE3D}). We chose to build upon EG3D, but our work is also applicable to other generators with a similar latent space.
For more information on 3D GAN architectures, we refer the interested reader to a recent survey~\cite{Xia20223DSurvey}.

\pgraph{Video Synthesis and Editing}
One branch of work attempts to leverage 2D GANs to generate video sequences ~\cite{Fox2021StyleVideoGAN, Tian2021Video, Yu2022Digan, Skorokhodov2022StyleGANV}. These ideas can be extended to create 3D videos~\cite{Bahmani20223DAware}, which also rely on 3D NeRFs.

\pgraph{GAN-based Video Editing}
GAN-based video editing is the core topic of this paper.
Duong \etal~\cite{Duong2019Aging} employ deep reinforcement learning for automatic face aging.
LatentTransformer~\cite{Yao2021LatentTransformer} encodes frames into the StyleGAN latent space using an encoder. They train a transformer to do attribute editing on single frames and blend the result with Poisson blending.
The main competitor to our work is Stitch it in Time (\SIT)~\cite{Tzaban2022STIT}, which crops the faces from a video, edits them with 2D GAN techniques, and merges the edited result back to the video with some blending. However, \SIT does not learn a 3D model of the human head, overfits to a particular video, and is unable to provide edits to the viewpoint of the human head.
Recently, VideoEditGAN (\VEG)~\cite{Xu2022VideoEditGAN} attempted to improve the temporal consistency of \SIT by running a two-step optimization approach focused on localized temporal coherence. 
Alaluf \etal~\cite{Alaluf2022Times} use StyleGAN3 for video editing, to leverage its inherent alignment capabilities and reduce texture sticking artifacts. 
Since this is an active area of research, all these techniques are concurrent work to our method, yet we do provide comparisons to showcase the benefits of our proposed approach.


\vspace{-1mm}
\section{Method}
\begin{figure}[t]
\centering
\includegraphics[width=\linewidth]{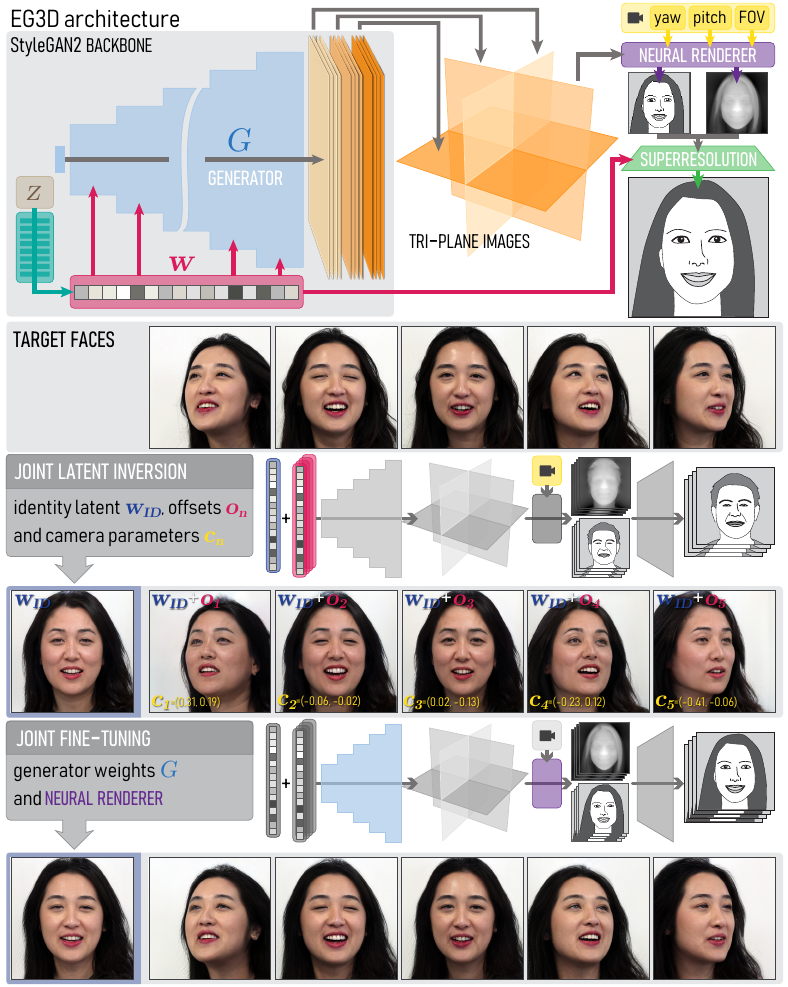}
   \vspace{-7mm}
   \caption{\textbf{Personalized Generator.} 
   First, we run a joint inversion on $N$ selected target faces, where we optimize a shared target person latent $\wperson$ and an offset $\offset{\idxS}$ for each face.
   This ensures the inversions share information about the target. Simultaneously, we jointly optimize for the camera pose $\cam{\idxS}$. 
   We then fine-tune the generator to ensure it captures the fine details of the target identity. 
   Note that the ``default'' latent \textit{(left column)} implicitly captures the identity of the target person without being explicitly optimized.}
   \label{fig:generator_inversion}
   \vspace{-5mm}
\end{figure}

\vspace{-1mm}
\label{sec:method}
In this section, we introduce the key components of \name to perform frame-by-frame video editing while allowing for rendering the edited face from new views. We leverage a 3D-aware generator that infers 3D geometry and camera positions while being trained solely on 2D images. We build a personalized 3D-aware generator by performing joint inversion on multiple frames and then use it to perform attribute editing, apply camera viewpoint changes, and finally composite the edited face rendered from a new view back into the original frame. An overview of our proposed approach is depicted in \cref{fig:pipeline} while the personalized generator architecture is shown in \cref{fig:generator_inversion}.

\subsection{Personalized 3D-Aware Generator}\label{sec:generator}
\pgraph{Face Selection and Cropping} To create a personalized 3D-aware GAN model, we start by processing a short range from the input video where \(N\) frames are selected such that they cover a range of orientations and facial expressions of the target person. We detect the facial keypoints within these frames using an off-the-shelf facial keypoint detector~\cite{Bulat2017Align} and use them to determine the face bounding box within the frame. This is achieved by calculating a rigid transformation from the facial keypoints in the frame to the facial keypoints in a generated example image, thereby aligning the keypoints at the center of the crop in the same way as the generator's original training data. We pick a specific field of view for cropping the faces and optimizing the generator, but the field of view remains a flexible parameter that can be adapted during any later stage in the pipeline.

\pgraph{Simultaneous Inversion} We propose to perform multiple inversions simultaneously.
EG3D has two major components in its generator. The first component uses a mapping network to map random vectors into a semantically meaningful space, called $\w$-space. Vectors in this space control a 3D neural field that defines a 3D representation that is rendered using volumetric rendering. The second component is a 2D upsampler that performs a \(4\times\) super-resolution on the original output.
We invert all selected faces simultaneously into the $\w$-space following a strategy similar to \cite{Roich2022PTI} that we discuss in detail below.

In order to find a representation in $\w$-space, we define a ``global'' \wperson aiming at capturing the global identity features of the target person, and a ``local'' offset vector \offset{\idxS} for each input expression \face{\idxS} that encodes the differences of each individual facial expression and position from the default \wperson.
The length of each \offset{\idxS} is regularized using an $\loss{L2}$ loss, aiming to keep the difference as small as possible, and capturing all similarities between the input images within the default person latent \wperson. 
We use a combination of a perceptual loss $\loss{LPIPS}$ and a pixel loss $\loss{L1}$ for the inversion. Note that during this stage, we calculate these losses on the raw output $\generator{\mtxt{raw}}(\wperson+\offset{\idxS})$ of the EG3D neural renderer at 128$\times$128 resolution because we observed that it yields sharper result quality rather than evaluating the loss at the output of the super-resolution network.
We downsample our target images to the same resolution \down{128}{\face{\idxS}} to compare. 
To ensure that we can faithfully capture the target person's identity and expression, we use BiSeNet~\cite{Yu2018BiSeNet} to obtain a segmentation \seg{exp}{\face{\idxS}} of the facial regions encoding the expression (eyes, mouth, eyebrows, and nose) and add an additional feature loss on this area to encourage consistent facial expressions (\eg closed eyes).
To obtain the inversion, in each optimization step, we sum up the losses for each face image \face{\idxS}, therefore jointly optimizing all targets simultaneously, yielding a total loss \loss{inv}.
\begin{align*}
\textstyle
\loss{inv} = \sum\nolimits_{\idxS=0}^{N}\weight{LPIPS} \, &\loss{LPIPS}(\generator{\mtxt{raw}}(\wperson+\offset{\idxS}), \down{128}{\face{\idxS}}) + \\
\weight{L1} \, &\loss{L1}(\generator{\mtxt{raw}}(\wperson+\offset{\idxS}), \down{128}{\face{\idxS}}) + \\
\weight{seg} \, &\loss{LPIPS}(\seg{exp}{\generator{}(\wperson+\offset{\idxS})}, \seg{exp}{\face{\idxS}}) + \\
\weight{reg} \, &\loss{L2}(\offset{\idxS})
\end{align*}
Due to the 3D awareness of the EG3D generator, the quality of the inversion into the latent space is highly sensitive to the camera parameter settings. Hence, in addition to optimizing for \wperson and \offset{\idxS}, we propose to also allow the inversion to optimize for the camera parameters \cam{\idxS} (\y{\idxS} and \p{\idxS}) for each input expression \face{\idxS}, which reliably estimates the camera position that the face is captured from.

A key advantage of this joint optimization is that the facial characteristics of the person preserve their high fidelity even when seen from novel views. When inverting a single image of a side-facing person into the EG3D latent space, exploring other viewpoints of the inverted latent can lead to significant distortions. Often, unseen features (\eg hidden ears) can be blurry or distorted, and the identity no longer resembles the input from a different viewpoint. The joint inversion, however, ensures that the different views are embedded closely enough in latent space such that even unseen views yield consistently identity-preserving outputs.

\pgraph{Generator Fine-tuning}  We propose a variant of Pivotal Tuning~\cite{Roich2022PTI} to jointly fine-tune the weights of the generator \generator{\mtxt{EG3D}} on all input faces \face{\idxS}, while keeping the detected \wperson, \offset{\idxS} and camera poses \cam{\idxS} fixed. Here, we do not allow the weights of the upsampler of the generator to be updated as we want to preserve the generalization capabilities of the super-resolution network and prevent it from overfitting to our target images. During this fine-tuning stage, we employ perceptual and pixel losses described as follows:
\begin{align*}
\textstyle
\loss{tune} = \sum\nolimits_{n=0}^{N}\weight{LPIPS} \, &\loss{LPIPS}(\gtuned(\wperson+\offset{\idxS}), \face{\idxS}) + \\
\weight{L1} \, &\loss{L1}(\gtuned(\wperson+\offset{\idxS}), \face{\idxS})
\end{align*}

Finally, we obtain a personalized EG3D generator \gtuned, fine-tuned to a set of facial expressions of the target person.
We verify that the fine-tuned generator indeed provides a good generalized latent space for the target person even though it was inverted and tuned based on a low number of frames by exploring the person created by the ``global'' latent code, which was not explicitly fine-tuned for, as well as through a latent space walk in the fine-tuned latent space.

\subsection{Frame-by-frame Video Inversion}
With the personalized 3D-aware generator in hand, we are now given a video of the same person as input which can be different from the one the generator was trained on.  
To process our new target video, we extract the facial keypoints from each frame \idxF to determine the location of the box to indicate the face crop within the frame. In order to stabilize the crop over time, which supports the temporal coherence of the inversion, we perform a Gaussian smoothing on the extracted facial keypoints along the temporal axis after extraction. However, it is important to not over-smooth, because fast motions in the video would yield distorted keypoint locations, deteriorating the inversion quality.

We then perform a frame-by-frame inversion of the extracted face regions \face{\idxF} into the space of the fine-tuned generator \gtuned. Like before, we optimize for an offset \offset{\idxF} for each input frame \face{\idxF}, as well as regularizing the offset length. After inverting the first frame, each consecutive frame \face{\idxF+1} is inverted starting from the previous inversion and only needs a low number (\mytilde50) of optimization steps. 
\vspace{-5mm}
\begin{align*}
\textstyle
\loss{vid} = \weight{LPIPS} \, &\loss{LPIPS}(\gtuned(\wperson+\offset{\idxF}), \face{\idxF}) +\\
             \weight{L1} \, &\loss{L1}(\gtuned(\wperson+\offset{\idxF}), \face{\idxF}) + \weight{reg} \, \loss{L2}(\offset{\idxF})
\end{align*}
This inversion yields a stack of offsets \offset{\idxF} from the identity latent \wperson as well as camera parameters \cam{\idxF} (\y{\idxF}, \p{\idxF}) encoding the expression and camera position for each frame.

\setlength{\belowcaptionskip}{-12pt}
\begin{figure}[t]
\centering
\includegraphics[width=\linewidth]{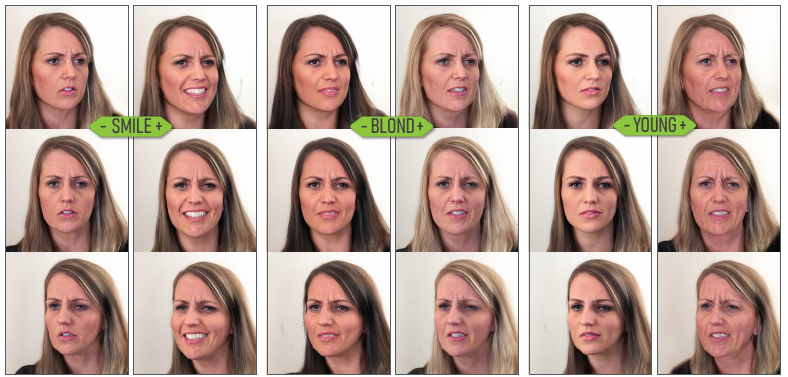} 
   \vspace{-6mm}
   \caption{\textbf{InterfaceGAN edits.} We show InterfaceGAN editing directions discovered in the latent space by applying them on our personalized generator. The attribute edits are consistent in 3D.
   }
   \label{fig:editing-directions}
\end{figure}

\subsection{Attribute Editing and Novel View Synthesis}
Since EG3D is built on top of StyleGAN2, we can leverage existing latent space editing techniques in order to discover semantic editing directions in the latent space of EG3D. As a proof of concept, we implemented InterfaceGAN~\cite{Shen2020InterFaceGAN} to find meaningful latent space direction vectors. We re-trained classifiers on the CelebA dataset for several facial attributes such as age, smile, gender, glasses, beard, and hair color and use these classifiers to classify the reference outputs of a set of randomized latent codes from our generator. Finally, we used an SVM to recover editing boundaries from these classified latent codes, which allows us to perform attribute editing in the EG3D latent space, as shown in \cref{fig:editing-directions}.
For a given latent space direction $\w_{\mtxt{dir}}$, we apply an edit as a linear combination of the person latent \wperson with the direction, multiplied by an empirically chosen weight $\alpha_{\mtxt{dir}}$, which can be positive or negative. For our video sequence, we use the edited person latent ${\wperson}'$ as the new identity to which we apply our video offsets $\offset{\idxF}$.
\begin{align*}
{\wperson}' = \wperson + \alpha_{\mtxt{dir}} \times \w_{\mtxt{dir}}
\end{align*}
In addition, we explore our temporal stack of latent encodings from novel views, diverging from the input views the inversion discovered. This allows us, to generate frontalized videos of the person by fixing the camera position or to define arbitrary camera trajectories around the subject.

\setlength{\belowcaptionskip}{-5pt}
\begin{figure}[t]
\centering
\includegraphics[width=\linewidth]{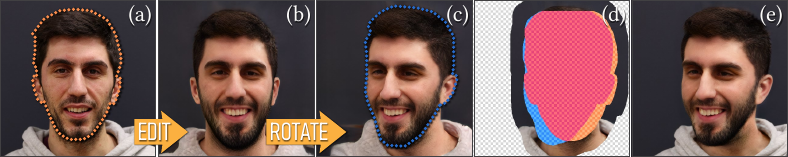}
   \vspace{-6mm}
   \caption{\textbf{Border Composition.} 
   We calculate the composition border based on face segmentations of the target image \figlabel{a} and the edited inset \figlabel{c}. We unite the masks and dilate the resulting joint mask to obtain a boundary around the face regions \figlabel{d} that should be optimized, which allows us to create the final composition \figlabel{e}. 
   }
   \label{fig:stitching-border}
   \vspace{-4mm}
\end{figure}

\subsection{Compositing with Source Video} 
After editing the inverted video, we want to recomposite the edited faces back with the source frames such that the edited video is temporally and spatially consistent. We accomplish this by running an optimization to ensure that the boundaries between the compositing of the edited face and the background of the source frame are smooth. 
Since cluttered video backgrounds are hard to reproduce consistently and without artifacts -- especially for novel views -- we define a compositing boundary region in a similar manner to \cite{Tzaban2022STIT}. To accomplish this, we need an accurate segmentation of the face and hair for both the original frame as well as the edited face. Hence, we use BiSeNet~\cite{Yu2018BiSeNet}, a semantic segmentation technique, that accurately provides such face semantics. We use the semantic regions for both the original and edited face to form a union of their respective masks, obtaining a boundary region around the relevant face region, as illustrated in \cref{fig:stitching-border}. 
We run a small number of optimization steps, optimizing for the boundary region of the edited image to appear as close as possible to the boundary region of the input frame while retaining the appearance of the edit. Finally, we use an affine transformation to re-insert the cropped face region back into the original frame and we alpha blend along the optimized boundary region to seamlessly composite the edit with the source frame.

\subsubsection{Flow-based View Adjustment}
\begin{figure}[t]
\centering
\includegraphics[width=\linewidth]{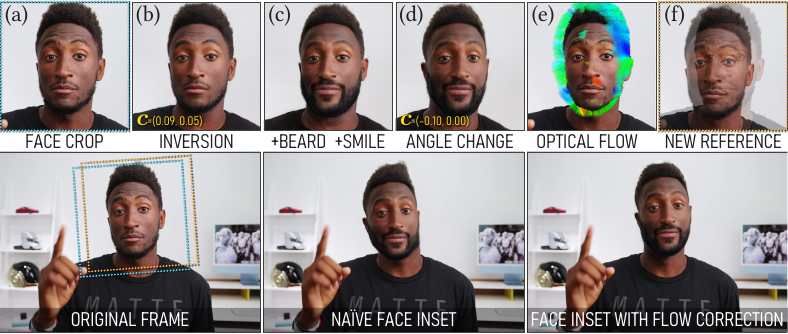}
   \vspace{-6mm}
   \caption{\textbf{View adjustment.} 
   After cropping \figlabel{a} and inverting \figlabel{b} a face, we perform face editing \figlabel{c} and change the camera viewpoint to an unseen angle \figlabel{d}. When replacing the face in the original frame with this edit, it yields poor quality \textit{(bottom center)} even for small angular changes, because the rotated face is in the wrong location with respect to the body. We address this by estimating the optical flow \figlabel{e} between the face crop and the edit and use the flow direction to correct the location of the reference face based on the prospective inset \figlabel{f}. This allows us to composite the edited face into the frame in a natural-looking fashion \textit{(bottom right)}.
   }
   \label{fig:camera-problem}
   \vspace{-4mm}
\end{figure}

During the process of re-inserting the edited face \face{\edit} back into the source frame \source, a major challenge arises when the camera pose has been modified and the face is rendered from a novel view. 
This is because the face is oriented within the bounding box based on the facial keypoints as described in \cref{sec:generator} while upon a camera pose change, the face pivots around the keypoints. When, for instance, attempting to replace a face viewed at an angle with a frontalized face while retaining the original crop boundary of the inset, the keypoints are still roughly in the same location, yet the mass of the head, and crucially, the neck is shifted according to the face rotation, as seen in \cref{fig:camera-problem}, which results in the inset being disconnected from the rest of the body even when using a boundary stitching technique. 

\setlength{\belowcaptionskip}{-5pt}

To alleviate this problem, we introduce a simple yet effective technique to reposition the reference face region within the source frame.
We discover the optimal position of the updated inset with respect to the source frame by estimating the optical flow between the face segmentation in the source frame \source and the face segmentation in the inset region \face{\edit} after camera rotation. We convert the images to grayscale and use Farneb\"{a}ck optical flow~\cite{Farnebaeck2003OpticalFlow} to evaluate a dense flow field of the displacement between the edited and target faces. 
The optical flow is defined as a 2D displacement vector field $\vec{d}$ with the displacement vector at image position $(x, y)$ given by \(\vec{d}(x, y) = (u(x, y), v(x, y))\)
where the correspondence between the two images \face{\idxF} and \face{\edit} is:
\begin{align*}
\face{\edit}( x + u(x, y), y + v(x, y) ) = \face{\idxF}( x,  y ).
\end{align*}
We then compute vector magnitudes $\|\vec{d}\| = \sqrt{ u^2+v^2}$ and directions $\phi=\mtxt{atan2}{(v, u)}$, respectively. After eliminating all vectors with a magnitude smaller than a threshold $\epsilon$, we create a histogram of all remaining  directions. We define a dominant displacement vector $\vec{d}_{\mtxt{dom}}$ from the median direction of the histogram bin with the largest count and the maximum vector length within that histogram bin. 
This ensures that erroneous flow directions from features that are present within one of the two images but not the other are not contributing to the final output.

The displacement vector $\vec{d}_{\mtxt{dom}}$ is reprojected from inset space into frame space and is used to correct the location of the reference face crop.
To ensure a smooth transition between adjacent frames, we perform temporal Gaussian smoothing of the recovered displacement vectors.
We then apply our inset optimization using the updated reference areas and obtain a significantly more faithful result, allowing for large camera changes with natural-looking results.


\begin{table}[t]
\arrayrulecolor{mediumgray}
\small
\centering
\setlength{\aboverulesep}{0pt}
\setlength{\belowrulesep}{0pt}
\caption{\textbf{Video Quality Metrics.} We compare the quality of our inversion with \SIT using reconstruction metrics on a subset of the \textit{VoxCeleb} dataset. We also evaluate the Fr\'echet Inception Distance~(FID) of inversion and edits with respect to the source video.}
\vspace{-0.7em}
\resizebox{\columnwidth}{!}{
\begin{tabular}{l|cc|ccc}
                 &\multicolumn{2}{c|}{\textbf{Reconstruction Quality}}&\mc{3}{\textbf{Editing Quality}}                                        \\
                 &\gr\textbf{PSNR}~$\uparrow$&\gr\textbf{SSIM}~$\uparrow$ & \mc{3}{\gr\textbf{FID}~$\downarrow$}                               \\    
    \tss{Method} &\multicolumn{2}{c|}{\tss{Inversion}}             & \tss{Inversion}         & \tss{Age Edit}          & \tss{Angle Edit}      \\
    \midrule
    \arrayrulecolor{lightgray}
       \SIT      &            38.0134  &            \textbf{0.9798}&                  6.8329 &                 15.8021 &             19.0371   \\
       \cc \name &\cc \textbf{38.1259} &\cc                 0.9704 & \cc      \textbf{4.9852}&\cc     \textbf{~~9.3410}&\cc   \textbf{~~8.9953}\\ 
    \bottomrule
\end{tabular}
}
\vspace{-5mm}
\label{tab:reconstruction}
\end{table}

\section{Experiments}
We conduct a wide range of quantitative and qualitative comparisons to demonstrate the key contributions of our work along with ablation studies against baselines and simplified variants where proposed modules are removed. We showcase that \name is on par with \SIT in terms of reconstruction quality while it also greatly outperforms prior work in editing quality and identity preservation. However, a key novelty afforded by \name is the ability to render the edited face from novel viewpoints within the existing frame, a task for which comparisons are hard due to the absence of ground truth. We showcase this with qualitative results and videos provided in the supplementary material.

\subsection{Quantitative Evaluation}\label{sec:evaluation}
We establish comparisons with \SIT~\cite{Tzaban2022STIT}, the key competitor to our work in the field of GAN video editing, and with \VEG~\cite{Xu2022VideoEditGAN} for which many components (\eg their stitching technique) are identical to \SIT, so we only compare facial similarity metrics. 
Please see the supplementary material for a detailed description of the comparison setup between our method and the competitors.

\pgraph{Inversion Quality} We provide a quantitative comparison of our inversion quality by measuring the reconstruction quality with respect to the input video in \Cref{tab:reconstruction}. We evaluate PSNR and SSIM for our method and for \SIT on a set of 16 videos from the \textit{VoxCeleb}~\cite{Nagrani2017VoxCeleb_OG} dataset, inverting the face region and recompositing it with the source video without edits. Both methods perform well on the reconstruction of the input signal and the final reconstruction quality of our technique is on par with \SIT.

\begin{table}[t]
\arrayrulecolor{mediumgray}
\footnotesize
\setlength{\tabcolsep}{4pt}
\setlength{\aboverulesep}{0pt}
\setlength{\belowrulesep}{0pt}
\centering
\caption{\textbf{Face Similarity Metrics.} We evaluate the identity preservation of inversion and edits based on the cosine similarity of Arc\-Face features extracted from generated face crops with respect to the face crops of the source video. To evaluate coherence over time, we measure the dissimilarity between consecutive frames.
}
\vspace{-0.8em}
\resizebox{\columnwidth}{!}{
\begin{tabular}{r|l|c|c}
                       &\tss{Method}&\gr{\textbf{Similarity to Source}}~$\uparrow$ &\gr{\textbf{Temporal Difference}}~$\downarrow$ \\
\midrule
\arrayrulecolor{lightgray}
 \mr{3}{Inversion}     & e4e        &           0.6923 &   6.2851             \\
                       & \SIT       &  \textbf{0.9261} &  1.2361              \\
                       & \cc \name  &       \cc 0.9203 &\cc \textbf{1.0444}   \\
  \cmidrule{1-4}
 \mr{3}{Age Editing}   & \SIT       &           0.7891 &    1.4126            \\
                       & \VEG       &           0.6004 &    1.7159            \\
                       & \cc \name  &\cc \textbf{0.8381}&\cc \textbf{1.2257}  \\
  \cmidrule{1-4}
 \mr{3}{Angle Editing} & \SIT       &            0.6695 &    1.3102           \\
                       & \VEG       &            0.4955 &    1.4502           \\
                       & \cc \name  &\cc \textbf{0.8694}&\cc \textbf{1.2761}  \\
\bottomrule
\end{tabular}
}
\vspace{-5mm}
\label{tab:face-similarity}
\end{table}

\pgraph{Image Quality} To evaluate the image quality produced by the respective techniques, we compute the Fr\'echet Inception Distance (FID), which is a commonly used quality metric for GANs. To obtain the FID for each video, we compare the set of all frames of the inverted video, as well as selected edits, with all frames of the source video. Our method is able to score very good FID scores overall (see \Cref{tab:reconstruction} right), confirming the quality of our results.

\pgraph{Face Fidelity} We calculate the fidelity of our inversion and edits based on a facial similarity metric, ArcFace~\cite{Deng2019ArcFace}, which extracts a 512-D feature vector capturing facial characteristics from a face region. We compute the metrics based on the respective face crops of the final inset region, scaled to 512$\times$512px, for our method, e4e encoding, \SIT, and \VEG in \Cref{tab:face-similarity} and average across a set of 16 videos from the \textit{VoxCeleb}~\cite{Nagrani2017VoxCeleb_OG} dataset.
The facial similarity is evaluated both with respect to the input video (frame-by-frame) as well as temporally by calculating the dissimilarity of adjacent frames in order to survey the temporal coherence of facial characteristics. In all cases, we use the cosine similarity between the extracted ArcFace deep features.
We observe that the quality of the inversion is good for both \SIT and \name, both significantly improving upon e4e encoding. \VEG uses the same PTI-based inversion as \SIT and is therefore not listed.
For latent space editing, our proposed \name leads the competition. \VEG exhibits good temporal coherence, however, the edits contain artifacts, resulting in a deterioration in the similarity to the source video. Additionally, our technique reconstructs the facial identity faithfully even for angle edits, a task in which \SIT and \VEG fail to produce plausible results due to their methods' inability to accommodate changes in the head rotation.

\begin{table}[b]
\arrayrulecolor{lightgray}
\scriptsize
\centering
\setlength{\aboverulesep}{0pt}
\setlength{\belowrulesep}{0pt}
\caption{\textbf{Timings and Memory Requirements.} We provide runtimes and Memory requirements for ours and competing methods.}
\vspace{-0.8em}
\begin{tabular}{r|c|c|c|c|c}
    \tss{Method}  & \gr{\textbf{Total}}      & \gr{\textbf{Precompute}} & \gr{\textbf{Main}}   & \gr{\textbf{Postprocess}} & \gr{\textbf{GPU}}  \\
    \toprule
    \SIT          & ~~58\mins53\secs         & ~~28\mins21\secs         & 30\mins19\secs       & ---                       & 22GB               \\
    \VEG          & 159\mins54\secs          & 107\mins~~9\secs         & 34\mins41\secs       &  18\mins~~3\secs          & 19GB               \\
    \cc\name      & \cc ~~35\mins43\secs     & \cc~~~~6\mins58\secs     & \cc 14\mins54\secs   & \cc 13\mins51\secs        & \cc 21GB           \\
\bottomrule
\end{tabular}
\label{tab:timings}
\vspace{-2.5mm}
\end{table}

\pgraph{Resource Usage} We compare runtimes and memory requirements for the default pipeline of related methods and our method, respectively, in \Cref{tab:timings}. We split each method, wherever applicable, into precomputation, main method, and postprocessing steps and used the hyperparameter settings according to the authors' suggestions. Note that the precomputation step in our method has to only be run once per identity and can then be applied to multiple videos. All experiments are performed on an example video consisting of 200 frames at a resolution of 1920$\times$1080px, using a single NVIDIA A100 GPU with 40GB memory.

\pgraph{Ablation Study} We quantitatively verify the effectiveness of different architecture choices we made, as shown in \Cref{tab:ablation}. We compare several metrics with respect to the source video for our default implementation, and the ablation experiments, respectively. We run four experiments on a set of 5 videos: (1) \name without generator fine-tuning, (2) \name without flow-based adjustment, (3) \name without the joint \wperson latent and offset regularization, (4) using only a single input face for the generator inversion and fine-tuning, which in practice is almost identical to only frame-by-frame inversion in EG3D without any personalized generator. We demonstrate that all experiments deteriorate the facial fidelity as well as the reconstruction quality.

\begin{table}[t]
\arrayrulecolor{lightgray}
\scriptsize
\centering
\setlength{\aboverulesep}{0pt}
\setlength{\belowrulesep}{0pt}
\caption{\textbf{Ablation Study.} We demonstrate the effect of removing various components of our pipeline on several quality and reconstruction metrics. We measure the face similarity using the cosine similarity of ArcFace features of the generated face crop, and the reconstruction metrics at the target video resolution.}
\vspace{-0.8em}
\resizebox{\columnwidth}{!}{
\begin{tabular}{l|c|c|c}
            \textbf{Ablation Type}&\gr\textbf{Face Similarity~$\uparrow$}& \gr\textbf{PSNR~$\uparrow$}   & \gr\textbf{SSIM~$\uparrow$}    \\ 
    \midrule
     \cc \name, full              & \cc      \textbf{0.9101}&\cc\textbf{35.5367}&\cc\textbf{0.9763} \\
     no generator fine-tuning     &                  0.7191 &           33.1202 &           0.9616  \\
     no flow correction           &                  0.8198 &           24.8845 &           0.9350  \\
     no regularization            &                  0.8007 &           25.6537 &           0.9162  \\
     single input for inversion   &                  0.7382 &           25.1950 &           0.9137  \\
     \bottomrule
\end{tabular}
}
\label{tab:ablation}
\vspace{-5mm}
\end{table}

\begin{figure}[b]
\centering
\vspace{-6mm}
\includegraphics[width=\linewidth]{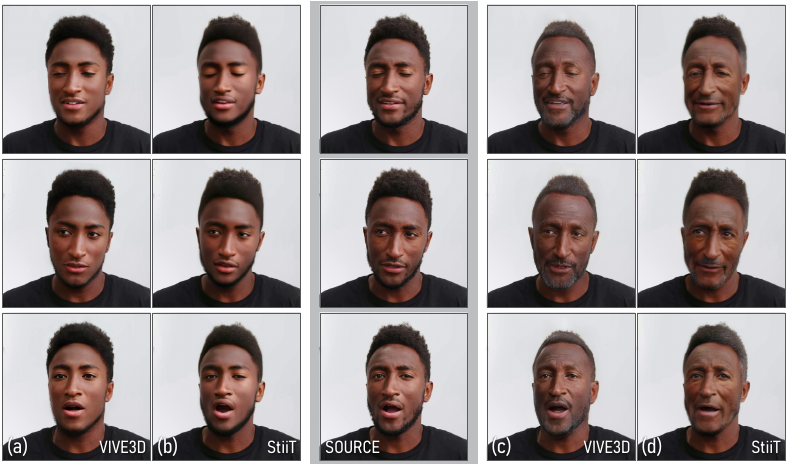} 
   \vspace{-6mm}
   \caption{\textbf{Comparisons with \SIT on attribute editing.} 
   We show an example of our method and \SIT editing the subject's age in the video frame \textit{(center column)}. Both methods yield plausible but distinctly different results. Our results \textit{(columns \figlabel{a}, age=--1.4 and \figlabel{c}, age=+2.3)} vs \SIT \textit{(columns \figlabel{b}, age=--8 and \figlabel{d}, age=+12)}. 
   }
   \label{fig:comparison-age}
\end{figure}

\label{sec:results}
\subsection{Qualitative Evaluation}
\pgraph{Semantic Edits} First, we demonstrate that the quality of semantic edits using \name, adapting well-established latent space editing techniques for 3D GANs, is on par with the editing quality of \SIT, as shown in \cref{fig:comparison-age}. Note that while both approaches discover latent space directions using InterfaceGAN~\cite{Shen2020InterFaceGAN}, the latent spaces and discovered directions are dissimilar, yet both plausible.

\setlength{\belowcaptionskip}{-5pt}
\begin{figure}[t]
\centering
\includegraphics[width=\linewidth]{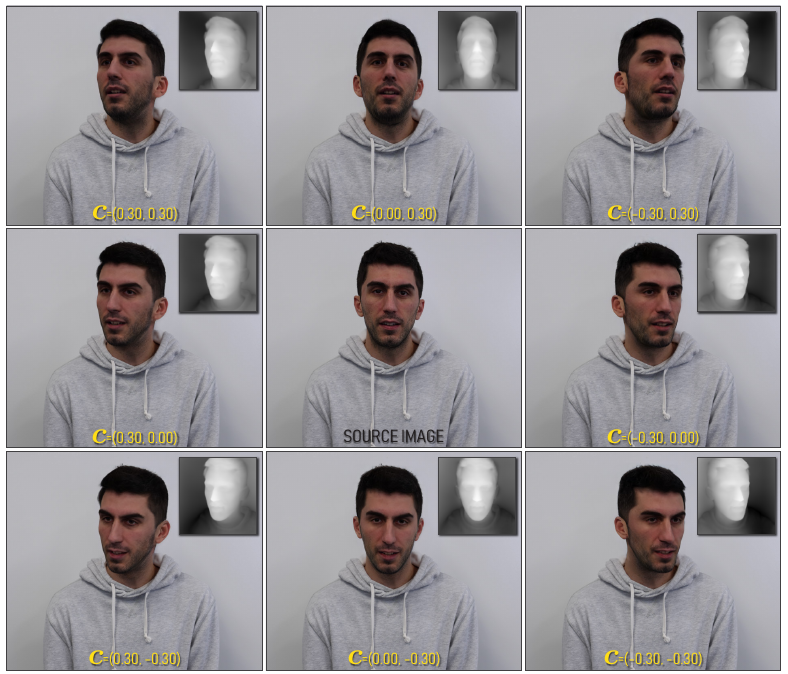} %
   \vspace{-6mm}
   \caption{\textbf{Changing camera poses.} Our method can freely alter the camera pose and composite the result back with the source frame by fixing the divergence between the source and target pose using our optical flow correction strategy. The generated results look natural despite the static body pose.
   }
   \label{fig:view-directions}
   \vspace{-2mm}
\end{figure}

\begin{figure}[t]
\centering
\includegraphics[width=\linewidth]{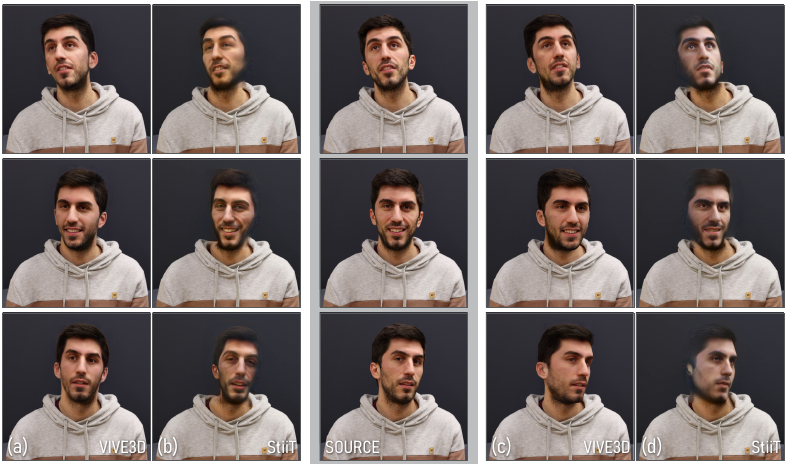}
   \vspace{-6mm}
   \caption{\textbf{Comparisons with \SIT on viewpoint changes.} 
   \name \textit{(\figlabel{a} and \figlabel{c})} produces plausible results for both positive and negative yaw changes, whereas \SIT \textit{(\figlabel{b} and \figlabel{d})} is unable to create natural compositions of the edit with the target frame.
   }
   \vspace{-0.3cm}
   \label{fig:comparison-fail}
\end{figure}

\pgraph{Synthesizing novel views} We show that \name can accommodate changes in the camera view with natural-looking results for a wide range of views regardless of the input face orientation, as shown in \cref{fig:view-directions}. This is nontrivial as we need to ensure that the person's identity is consistent from multiple viewpoints while the body in the source frame also defines a rigid constraint to which the head alignment must be adjusted to.
While \SIT produces good results for attribute edits, it cannot composite images with angular changes, despite the fact that a limited head pose change can be achieved by applying latent space manipulations. In \cref{fig:comparison-fail}, we show a comparison where \SIT is unable to generate a reasonable composition, whereas we can achieve natural-looking results for a similar head rotation.

\begin{figure}[t]
\centering
\includegraphics[width=\linewidth]{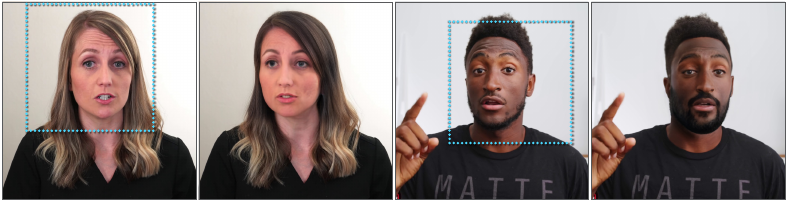} 
   \vspace{-6mm}
   \caption{\textbf{Spatial Consistency.} \name composites images with challenging boundaries such as long hair \textit{(right)}, yielding faithful hair color change results. For hard boundary cases, such as matching with a static piece of hair outside the boundary crop \textit{(left)}, it plausibly connects the contents of the two images.}
   \label{fig:stitch-boundaries}
   \vspace{-2mm}
\end{figure}

\pgraph{Compositing with challenging boundaries} When parts of the head or hair are visible both inside and outside the face bounding box then compositing the edited frame back into the original input is a challenging task. In most cases, we address these scenarios by adjusting our camera parameters (\eg, choosing a wider field of view for our generator) but some configurations can still be challenging, especially when different textures need to be matched. We show in \cref{fig:stitch-boundaries} that our technique attempts to produce plausible inset optimizations even for such instances.

\begin{figure}[t]
\centering
\includegraphics[width=\linewidth]{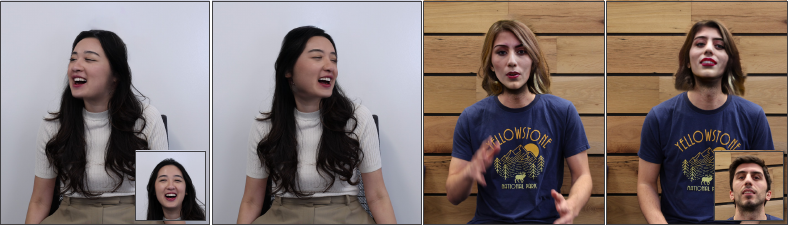}
   \vspace{-6mm}
   \caption{\textbf{Limitations.} 
   \name inherits some limitations from the frameworks we rely on. EG3D cannot capture extreme poses well \textit{(left)}. Large angle changes cannot be composited naturally with the static body in the source frame. For extreme edits (gender edits that change the hair structure \textit{(right)}), it is difficult to yield temporally consistent results, both due to the entanglement of the latent space editing and the challenges of frame compositing.
   }
   \vspace{-0.3cm}
   \label{fig:limitations}
\end{figure}

\pgraph{Limitations} Changing the camera parameters for the head in videos with fast motion or discontinuities results in artifacts because the flow estimation becomes unstable. Furthermore, we inherit the shortcomings of EG3D (see \cref{fig:limitations}): stronger entanglement of attribute edits compared to StyleGAN2, extreme camera poses are not captured in the training set, and texture sticking. Finally, since the video outside the face region is immovable, the range of possible changes is constricted to plausible compositions.

\vspace{-0.2cm}
\section{Conclusion}\label{sec:conclusion}

In this paper, we introduced \name, a novel framework that uses prior information encoded in 3D GANs for video editing. Our edits are identity-preserving and temporally consistent. While we enable standard semantic edits, such as age, or expressions, a distinguishing feature of our work is that we facilitate edits that alter the view of the head. This capability is not available in any 2D GAN-based prior work. %
The key building blocks of our work are a new embedding algorithm that jointly embeds multiple frames and optimizes for camera pose as well as flow-guided video compositing.
In future work, we aim to extend our framework to include a 3D GAN for head details and another 3D GAN for the body. We also plan to investigate performance speedups by replacing various optimizations with encoders.

{
\small
\bibliographystyle{ieee_fullname}
\bibliography{bibliography}
}
\end{document}


\title{\name: Viewpoint-Independent Video Editing using 3D-Aware GANs \\ \textsc{Supplementary Materials}}
\author{
    Anna Fr\"{u}hst\"{u}ck\textsuperscript{{2}}, 
    Nikolaos Sarafianos\textsuperscript{1},
    Yuanlu Xu\textsuperscript{1}, 
    Peter Wonka\textsuperscript{2}, 
    Tony Tung\textsuperscript{1}
\\[0.6ex]
	\textsuperscript{1~}Meta Reality Labs Research, Sausalito\quad
        \textsuperscript{2~}KAUST \quad \\ 
        \small{\tt\href{https://afruehstueck.github.io/vive3D}{afruehstueck.github.io/vive3D}}
}

\twocolumn[{%
\renewcommand\twocolumn[1][]{#1}%
\maketitle
\vspace*{-.5in}
\begin{center}
  \centering
  \captionsetup{type=figure}
    \includegraphics[width=\linewidth]{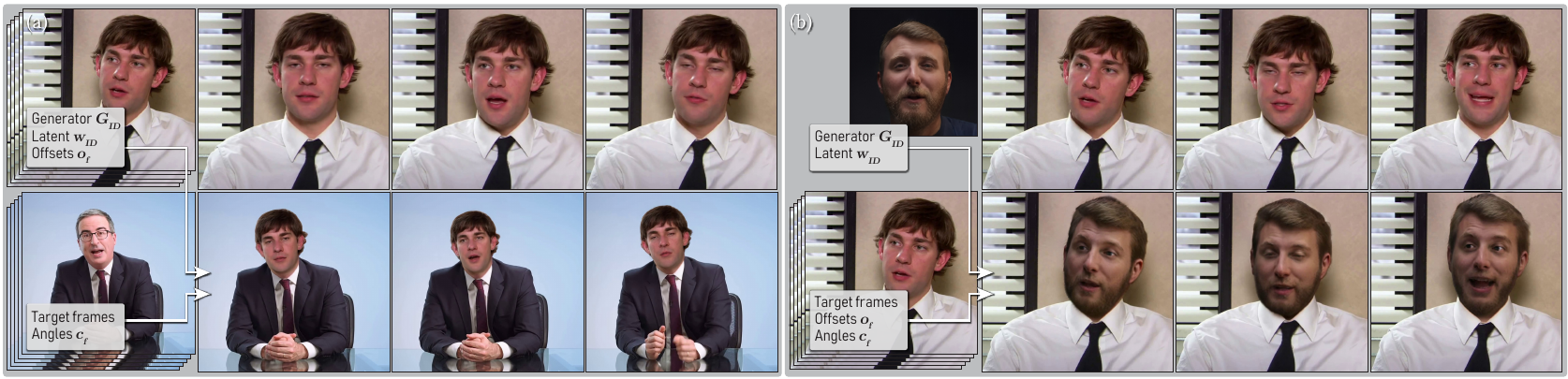}
   \caption{\textbf{\name generalization to new identities}. The benefits from decomposing the inversion of the input into an identity latent and a set of offsets unlock applications of face/motion re-targeting with minimal effort. This is possible due to our novel personalized generator that can be trained on a specific person's identity and then applied to edit an unseen video. We show two examples: In example \figlabel{a}, we use a person \textit{(top row)} to invert and fine-tune the generator, and we determine the video offsets based on this video sequence. The bottom row determines the target frames as well as the face location and angles. For example \figlabel{b}, we use a personalized generator \textit{(top left)}, but the target frames, angles, as well as motion, stem from a distinct video, driving the motion of the target person.
    }
   \label{fig:head-change}
\end{center}%
}]

\begin{figure*}[t]
\centering

\end{figure*}

\section{Additional Results}
\subsection{Supplementary Video}
Please see our supplementary video (on the \href{https://afruehstueck.github.io/vive3D}{project webpage}) for video sequences illustrating our proposed method and a set of results demonstrating the unique capabilities of our technique as well as comparisons to related methods.

\begin{figure*}[t]
\centering
\includegraphics[width=\linewidth]{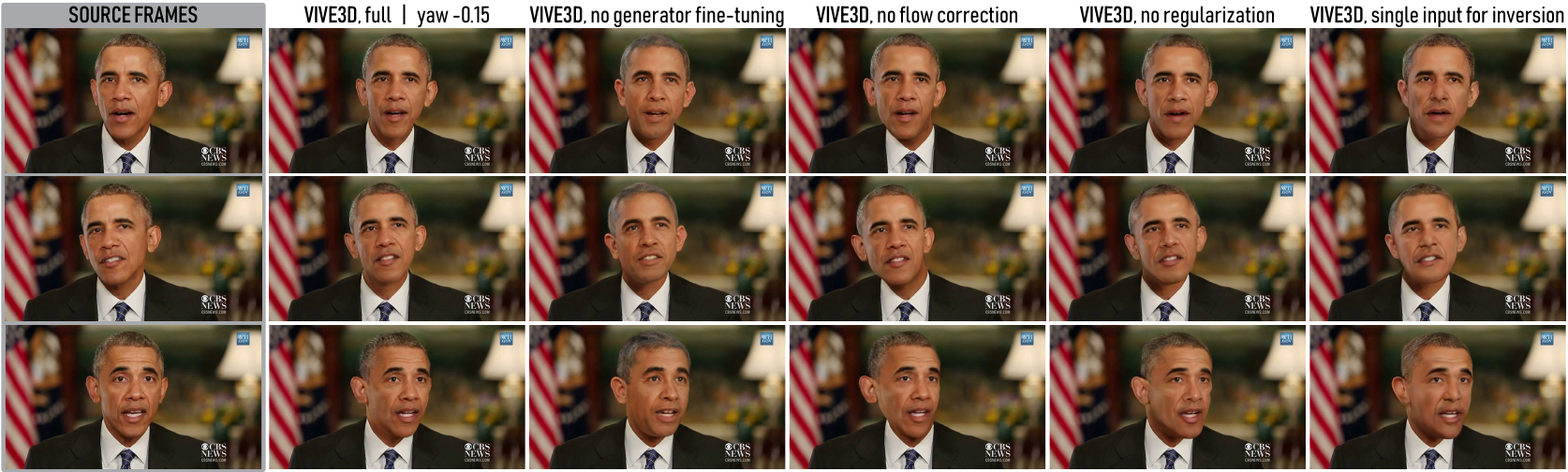}
   \caption{\textbf{\name Ablation Study.} Our proposed approach \name \textit{(\(2^{nd}\) column)} with all the proposed components demonstrates better identity preservation, fixes the spatial misalignment between the face and neck when rendered from a novel view and better captures the fine-level details of the face and results in high-fidelity faithful renders of the person from new views. 
    }
   \label{fig:ablation-images}
   \vspace{-1.5em}
\end{figure*}

\subsection{Experimental Edits}
We showcase some additional experimental edits to illustrate the generalization abilities of our approach for general-purpose video editing. For example, we are able to use two completely disjoint videos of different subjects and achieve reasonable results at compositing them.
We show in \Cref{fig:head-change} two instances of such applications: On the left (\Cref{fig:head-change} (a)), we use one personalized Generator with its "default" person latent \wperson, and a stack of video offsets encoding a sequence of face motions. We then compose these with a different body by running our inset optimization, using the target video frames and head angles from that particular video.
On the right (\Cref{fig:head-change} (b)), we use the encoded face motion from the target video, projecting the motion onto a different person's face, thereby essentially replacing the head in the target video. 
Note that in order to achieve plausible results for these instances, we need to copy head and neck due to the slight differences in lighting between the source and target faces. Further, the segmentation masks need to be considered carefully and be big enough, \eg in the right result, the hair sticking out from the source person's head needs to be covered by inpainting with background color during the inset optimization in order to achieve a reasonable result. Otherwise, the optimization will add extra hair and change the hairstyle.
We observe that \name generates realistic results of placing the first person's head on the second person's body that are temporally and spatially consistent and follow the target frame motion. This is possible because of our proposed design (personalized generator, separate identity, and offset latents) which makes \name unique compared to prior work. It is worth noting that when the source and target videos have different light conditions then the results might have different lighting between the body and the face. 
This is because we do not explicitly tackle this problem in our architecture and hence lighting is baked in the final generated/edited head before it is placed on top of the new body. We identify the problem of better lighting transfer as an avenue for future work.

\subsection{Ablation Study}
To evaluate the impact of each module we conduct ablation studies and report our quantitative and qualitative results in Table 4 of the main paper and \Cref{fig:ablation-images} of the supplementary respectively. Given a video of a person talking \textit{(\(1^{st}\) column) }, we demonstrate our complete approach when rendering the output video from a new viewpoint \textit{(\(1^{st}\) column)}. 
In the next 4 columns of results, we strip one component at a time and observe different performance quality drops. For example, if we do not fine-tune the generator it is clear that the identity of the individual is not properly preserved \textit{(\(3^{rd}\) column)}. If we remove the flow correction module which is a key contribution of our approach, we observe that the face and the neck are not well aligned which makes the results seem unnatural \textit{(\(4^{th}\) column)}. The impact of the flow correction module is demonstrated also in \Cref{fig:camera-problem_sup} and discussed in detail in the supplementary video. 
If we strip the regularization \textit{(\(5^{th}\) column)}, we remove the joint latent \wperson and treat each target face in the initial inversion separately. This means we don't constrain the individual latents to stay close to the common latent. We can see that this leads to a deterioration in the inversion quality of the video frames and produced artifacts, as the inverted latents no longer share information, and there is no constraint on the projected location in latent space. Finally, if we were to only perform single-frame inversion rather than multi-frame, we also observe a significant drop in fidelity \textit{(\(6^{th}\) column)}, which indicates that the proposed approach of performing multi-frame fidelity is beneficial as it better captures the identity and the fine details of the face.

\begin{figure}[t]
\centering
\includegraphics[width=\linewidth]{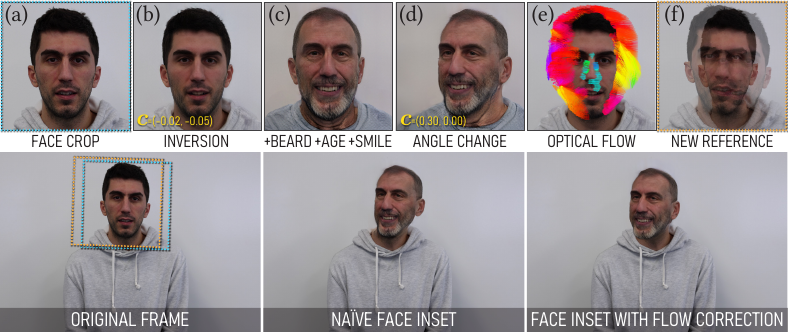}
   \vspace{-6mm}
   \caption{\textbf{View Adjustment Additional Results.} 
   We illustrate the problem arising when attempting to composite a changed view of a person's head on top of the original body. 
   After cropping \figlabel{a} and inversion \figlabel{b}, we perform face editing \figlabel{c} and change the camera viewpoint to an unseen angle \figlabel{d}. Replacing the face in the original frame with this edit yields poor quality \textit{(bottom center)} even for small angular changes because the rotated face is in the wrong location with respect to the body. We address this by estimating the optical flow \figlabel{e} between the face crop and the edit and use the flow direction to correct the location of the reference face based on the prospective inset \figlabel{f}. This allows us to composite the edited face into the frame in a natural-looking fashion \textit{(bottom right)}.
   }
   \label{fig:camera-problem_sup}
   \vspace{-4mm}
\end{figure}

\subsection{Qualitative Results of Method Comparisons}
We provide a qualitative comparison to related methods of GAN-based video editing, Stitch it in Time (\SIT)~\cite{Tzaban2022STIT} and VideoEditGAN (\VEG)~\cite{Xu2022VideoEditGAN}. 
First, we discuss some of the differences between our proposed method and the related work to establish the parameters of our comparison.
\begin{figure}[t]
\centering
\includegraphics[width=\linewidth]{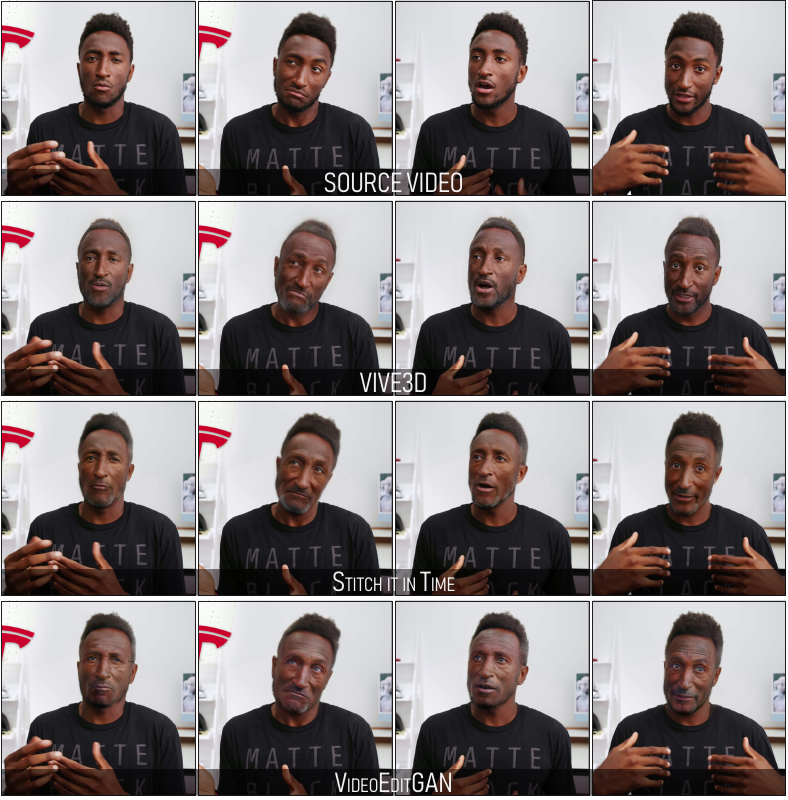}
    \vspace{-1.6em}
   \caption{\textbf{Comparison to related work on age editing.} Due to the distinctly different InterfaceGAN editing directions, we handpicked the edit strength for the respective latent space edits to showcase a similar effect in aging the target person. Our technique yields results that are at least on par with the previous StyleGAN2-based editing techniques \textit{(bottom two rows)}. 
   \label{fig:methods-age-comparison_sup}
   }
   \vspace{-1.3em}
\end{figure}

\begin{itemize}[leftmargin=*]
\itemsep0em 
    \item \SIT and \VEG, which is closely related to \SIT, both use a StyleGAN2 backbone which outputs high quality images at 1024$\times$1024px resolution, whereas our backbone's (EG3D) output resolution is 512$\times$512px (obtained by super-resolution given 128$\times$128px inputs), providing 3D-awareness at the expense of slightly inferior image quality to classic StyleGAN2. In order to compare quantitatively, we downsample all results generated by \SIT and \VEG to 512$\times$512px, unless we compare at the full video resolution, in which case the output of the respective generator is already resampled to fit the resolution of the original face crop in the video frame.
    \item \SIT and \VEG rely on prior work for a reliable encoding framework, e4e~\cite{richardson2021encoding}, to yield good and coherent inversions, whereas we implemented an optimization strategy to obtain per-frame inversions. Implementing an encoding strategy for EG3D was outside the scope of our project, but would be an interesting topic for future endeavors. We expect that using an encoder would lower the embedding quality, but improve computation speeds.
    \item \SIT and \VEG fine-tune their generator on all video frames simultaneously, thus achieving very good coherence to the input. In contrast, we fine-tune on a select few target faces, yielding a more generalizable generator, which, in consequence, is not optimized to replicate the video frame-by-frame. 
    \item Like our approach, \SIT relies on InterfaceGAN~\cite{Shen2020InterFaceGAN} for discovering and applying latent space editing directions for many of their results. \VEG shows their results using edits based on StyleClip~\cite{Patashnik2021StyleClip}. To compare with their method, we adapted their code to also allow edits with InterfaceGAN -- analogous to \SIT -- before applying their temporal consistency strategy. Note that the discovered directions in StyleGAN2 and EG3D latent space do not yield identical results for the same attribute type and the strength needed to apply the direction vectors is different. We empirically chose weights to approximate the same edit strength when comparing results.
\end{itemize}

\begin{figure}[t]
\centering
\includegraphics[width=\linewidth]{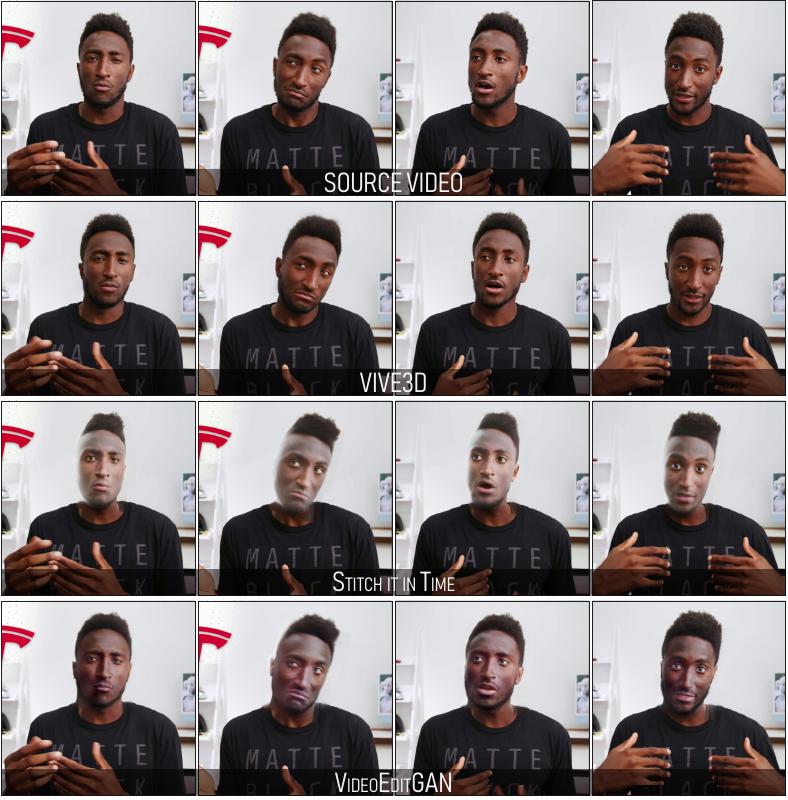}
    \vspace{-1.6em}
   \caption{\textbf{Comparison to related work on angle editing.} A slight change in angle is achieved for the related methods by applying a yaw-changing latent space direction. Both methods fail at producing a reasonable composition given these edits \textit{(bottom two rows)}. We implement a similar angle change and demonstrate that our method creates a natural-looking composition. 
   }
   \label{fig:methods-angle-comparison}
   \vspace{-1.3em}
\end{figure}

\begin{figure}[b]
  \captionsetup{type=figure}
  \centering
  \vspace{-0.6em}
  \caption{\textbf{Comparison to another 3D inversion technique.} To demonstrate the effectiveness of our Personalized Generator inversion, we compare the quality of our inversion with a recent technique of 3D GAN inversion~\cite{Ko20233D}. Note how the head shape and identity correspondence deteriorate when rotating the head away from the original pose.}
  \vspace{-0.6em}
  \includegraphics[width=\linewidth]{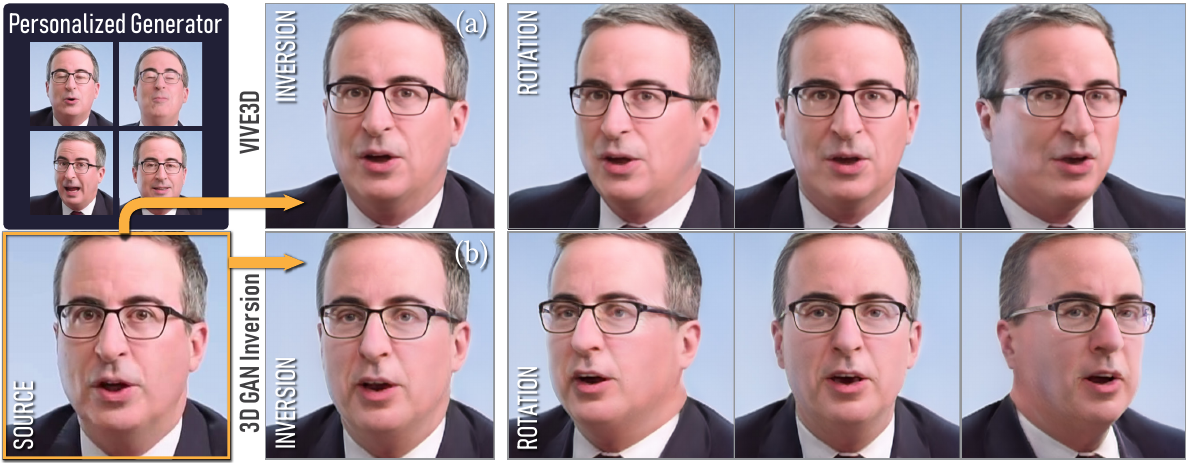}
  \label{fig:inversion-comparison}
   \vspace{-1.3em}
\end{figure}

We demonstrate in \Cref{fig:methods-age-comparison_sup} that all methods provide plausible results for classic semantic editing problems such as aging the target person. However, both related methods fail to yield plausible results for angle editing. In order to compare this task, we utilize a latent space direction discovered in StyleGAN2 that allows for slight angle changes. The comparison for these strategies is illustrated in \Cref{fig:methods-angle-comparison}.
The artifacts present in these qualitative results mirror the deterioration in quantitative scores indicated in Table 2 in the main paper.

In \Cref{fig:inversion-comparison}, we show a result of our multi-target inversion strategy (row \figlabel{(a)}) compared to another single-image 3D GAN inversion\cite{Ko20233D} (row \figlabel{(b)}). Please zoom in to observe the degradation of head shape and loss of identity when the head rotation diverges from the source image.

\begin{table}[t]
\arrayrulecolor{mediumgray}
\footnotesize
\setlength{\tabcolsep}{4pt}
\setlength{\aboverulesep}{0pt}
\setlength{\belowrulesep}{0pt}
\centering
\caption{\textbf{Reconstruction Metrics.} We compare the quality of our inversion with \SIT using reconstruction metrics on a subset of videos. We also evaluate the Fr\'echet Inception Distance~(FID) of inversion and edits with respect to the source video.}
\resizebox{\columnwidth}{!}{
\arrayrulecolor{mediumgray}
\setlength{\aboverulesep}{0pt}
\setlength{\belowrulesep}{0pt}
\begin{tabular}{l|l|cc|ccc}
      \multicolumn{2}{c|}{~}          &\multicolumn{2}{c|}{\gr\textbf{Reconstruction Quality}}&\mc{3}{\gr\textbf{Editing Quality}}                                     \\
      \multicolumn{2}{c|}{~}          &\textbf{PSNR}~$\uparrow$&\textbf{SSIM}~$\uparrow$ & \mc{3}{\textbf{Fr\'echet Inception Distance (FID)}~$\downarrow$}            \\    
                      &\textbf{Method}&\multicolumn{2}{c|}{\tss{Inversion}}              & \tss{Inversion}         & \tss{Age Edit}          & \tss{Angle Edit}        \\
    \midrule
    \arrayrulecolor{lightgray}
    \mr{2}{\textbf{Marques}}
                          & \SIT      &        \textbf{36.477}  &                  0.965 &                  10.11 &                   18.44 &                 21.58 \\
                          & \cc \name &\cc             33.791   &\cc       \textbf{0.987}& \cc     \textbf{~~7.25}&\cc        \textbf{12.73}&\cc      \textbf{11.63}\\  
    \cmidrule{1-7}
    \mr{2}{\textbf{Obama}}
                          & \SIT      &                34.969  &           \textbf{0.976} &        \textbf{~~3.80}&                   16.49 &                 17.12 \\
                          & \cc \name &\cc     \textbf{36.282} &\cc                0.969  & \cc            ~~3.88 &\cc       \textbf{~~8.67}&\cc     \textbf{~~7.22}\\ 
    \cmidrule{1-7}
    \mr{2}{\textbf{Dennis}}
                          & \SIT      &                40.708  &           \textbf{0.993} &                ~~6.32 &                   12.88 &                 16.96 \\
                          & \cc \name &\cc     \textbf{40.804} &\cc                0.990  & \cc    \textbf{~~4.07}&\cc       \textbf{~~8.36}&\cc     \textbf{~~8.07}\\
    \arrayrulecolor{mediumgray}
    \bottomrule
\end{tabular}
}
\label{tab:face-metrics-suppl}
\end{table}

\begin{figure}[t]
\centering
\includegraphics[width=\linewidth]{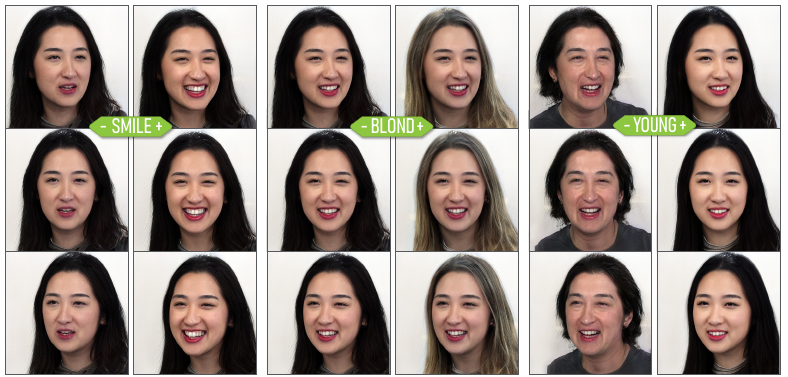}
   \caption{\textbf{InterfaceGAN Edits Additional Results.} 
   We show InterfaceGAN editing directions discovered in the latent space of the pretrained 3D GAN by applying them to our personalized generator. The attribute edits are plausible and consistent in 3D.
   }
   \label{fig:editing-directions_sup}
\end{figure}

\subsection{Additional Quantitative and Qualitative Results}
We provide some additional quantitative metrics on individual videos shown throughout this supplementary material. In \Cref{tab:face-metrics-suppl}, we analyze the reconstruction quality and editing quality for three individual videos, showing that our reconstruction and editing capabilities are on par with our main competitor technique \SIT for video inversion and editing tasks. We also analyze the inversion and editing performance of VIVE3D and related methods for two distinct videos in more detail in \Cref{tab:face-similarity-suppl}. We use the ArcFace~\cite{Deng2019ArcFace} metric to calculate the minimum, maximum, and average similarity to the source video as well as the temporal difference by evaluating the metric on adjacent video frames. The quantitative scores show that our method is superior for both attribute and angular edits, the latter being a task at which the previous 2D-GAN-based methods fail.

\begin{table}[t]
\arrayrulecolor{mediumgray}
\scriptsize
\setlength{\tabcolsep}{4pt}
\setlength{\aboverulesep}{0pt}
\setlength{\belowrulesep}{0pt}
\centering
\caption{\textbf{Face Similarity Metrics.} We evaluate the identity preservation of inversion and edits based on the cosine similarity of ArcFace features extracted from generated face crops with respect to the face crops of the source video. To evaluate coherence over time, we measure the dissimilarity between consecutive frames.
}
\vspace{-1em}
\resizebox{\columnwidth}{!}{
    \begin{tabular}{@{}c|r|l|ccc|ccc}
    \toprule         
     &                 &               &\multicolumn{3}{c|}{\gr\textbf{Similarity to Source}~$\uparrow$}& \mc{3}{\gr\textbf{Temporal Diff}~$\downarrow$}     \\   
     &                 &\textbf{Method}&\tss{min} &\tss{max} &        \tss{mean} &\tss{min}&\tss{max}&      \tss{mean} \\
    \midrule
    \arrayrulecolor{lightgray}
      \mr{9}{\rotatebox{90}{\textbf{Marques}}}
     & \mr{3}{Inversion}  & e4e        &    0.487 &    0.816 &            0.663  &    0.1 &    36.3 &            5.6   \\
     &                    & \SIT       &    0.720 &    0.877 &            0.820  &    0.1 &   ~~8.6 &    \textbf{1.3}  \\
     &                    & \cc \name  &\cc 0.820 &\cc 0.932 &\cc \textbf{0.894} &\cc 0.1 &\cc~~7.5 &\cc         1.4   \\
      \cmidrule{2-9}
     & \mr{3}{Age Edit}   & \SIT       &    0.720 &    0.877 &            0.820  &    0.1 &   ~~8.8 &            1.3   \\
     &                    & \VEG       &    0.430 &    0.668 &            0.551  &    0.1 &    11.5 &            1.9   \\
     &                    & \cc \name  &\cc 0.730 &\cc 0.923 &\cc \textbf{0.857} &\cc 0.1 &\cc~~5.2 &\cc \textbf{1.1}  \\
      \cmidrule{2-9}
     & \mr{3}{Angle Edit} & \SIT       &    0.654 &    0.801 &            0.740  &    0.1 &   ~~8.9 &            1.4   \\
     &                    & \VEG       &    0.424 &    0.685 &            0.568  &    0.1 &    12.1 &            1.7   \\
     &                    & \cc \name  &\cc 0.762 &\cc 0.899 &\cc \textbf{0.849} &\cc 0.1 &\cc~~8.1 &\cc \textbf{1.1}  \\
      \midrule
      \mr{9}{\rotatebox{90}{\textbf{Obama}}} 
     & \mr{3}{Inversion}  & e4e        &    0.469 &    0.801 &            0.665  &    0.1 &    44.8 &            5.9   \\
     &                    & \SIT       &    0.935 &    0.982 &    \textbf{0.968} &    0.0 &   ~~7.6 &            1.0   \\
     &                    & \cc \name  &\cc 0.882 &\cc 0.961 &\cc         0.930  &\cc 0.0 &\cc~~2.1 &\cc \textbf{0.5}  \\
      \cmidrule{2-9}
     & \mr{3}{Age Edit}   & \SIT       &    0.717 &    0.862 &            0.781  &    0.1 &   ~~6.1 &            1.0   \\
     &                    & \VEG       &    0.522 &    0.763 &            0.671  &    0.2 &   ~~7.1 &            1.5   \\
     &                    & \cc \name  &\cc 0.758 &\cc 0.903 &\cc \textbf{0.850} &\cc 0.1 &\cc~~4.0 &\cc \textbf{0.8}  \\
      \cmidrule{2-9}
     & \mr{3}{Angle Edit} & \SIT       &    0.668 &    0.827 &            0.753  &    0.0 &   ~~7.8 &            1.3   \\
     &                    & \VEG       &    0.771 &    0.874 &            0.840  &    1.0 &   ~~5.9 &            1.2   \\
     &                    & \cc \name  &\cc 0.782 &\cc 0.916 &\cc \textbf{0.868} &\cc 0.1 &\cc~~4.2 &\cc \textbf{0.9}  \\
    \arrayrulecolor{mediumgray}
    \bottomrule
    \end{tabular}
}
\vspace{-0.4cm}
\label{tab:face-similarity-suppl}
\end{table}

\begin{figure}[b]
\centering
\includegraphics[width=\linewidth]{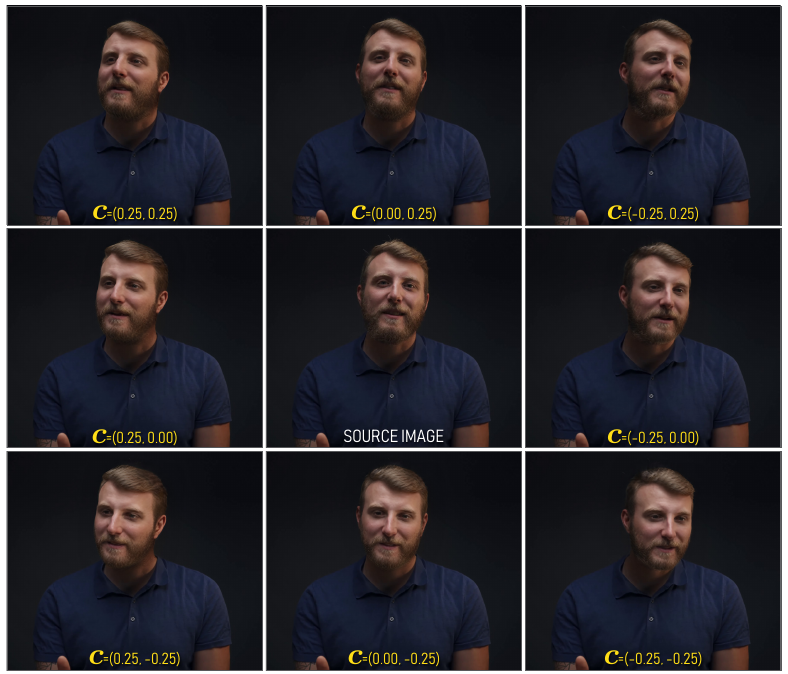}
   \vspace{-6mm}
   \caption{\textbf{Changing Camera Poses Additional Results.} Our method can freely alter the camera pose and composite the result back with the source frame by fixing the divergence between the source and target pose using our optical flow correction. The generated results look natural despite the static body pose.
   }
   \label{fig:view-directions_sup}
\end{figure}

\begin{figure*}[t]
  \captionsetup{type=figure}
  \centering
  \caption{\textbf{Additional Qualitative Results.} Given a video sequence, we process individual frames, cropping the face region to correspond with our generator's field of view. VIVE3D faithfully modifies several facial attributes as well as the camera viewpoint of the head crop. Finally, we seamlessly composite the edited face with the source frame in a temporally and spatially consistent manner, while retaining a plausible composition with the static components of the frame outside of the generator's region. The dotted squares in the center frame denote the reference regions for the three different camera poses in the column below.}
  \vspace{-0.5em}
  \includegraphics[width=\linewidth]{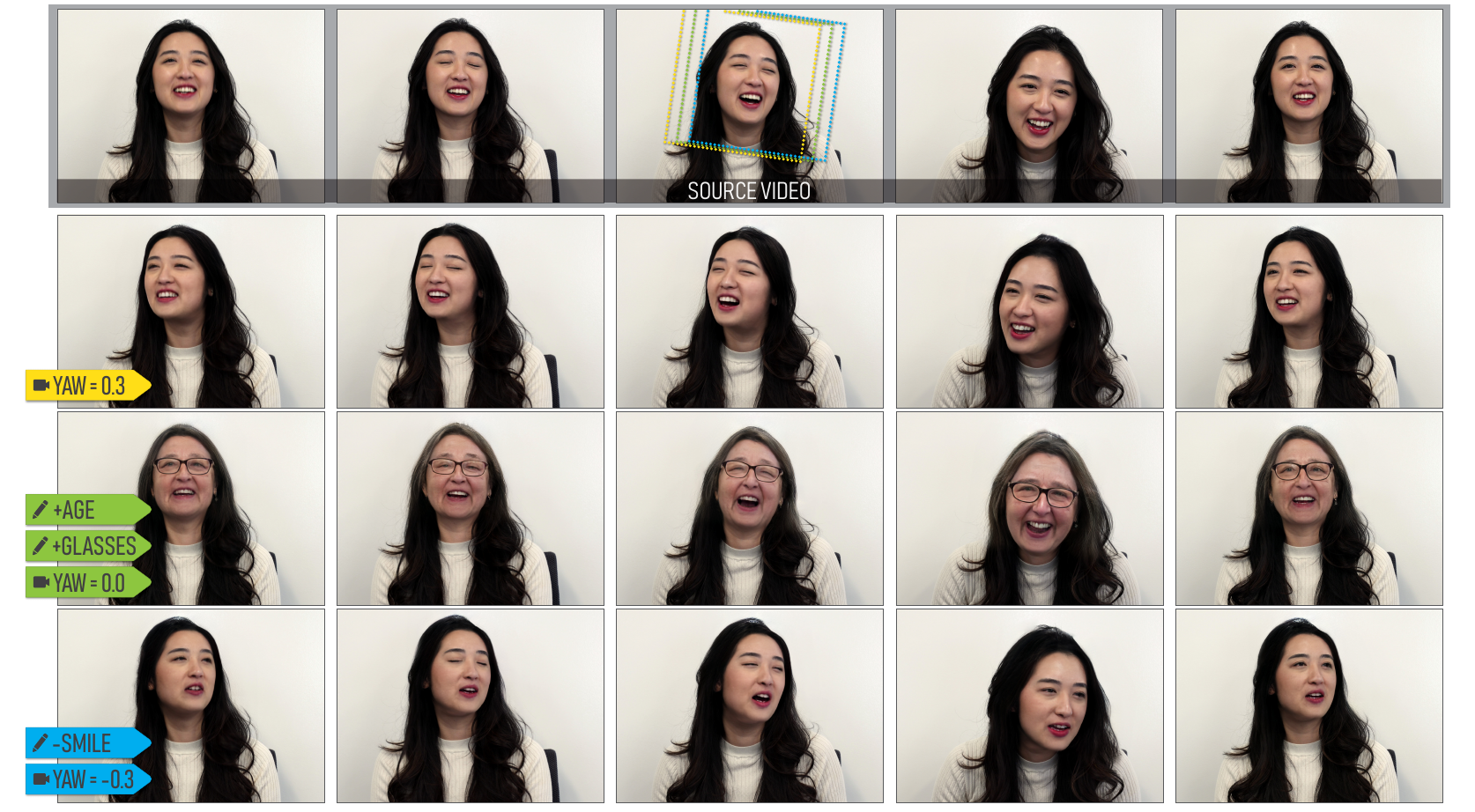} 
  \label{fig:teaser_sup}
  \vspace{-1em}
\end{figure*}

\begin{figure}[t]
\centering
\includegraphics[width=\linewidth]{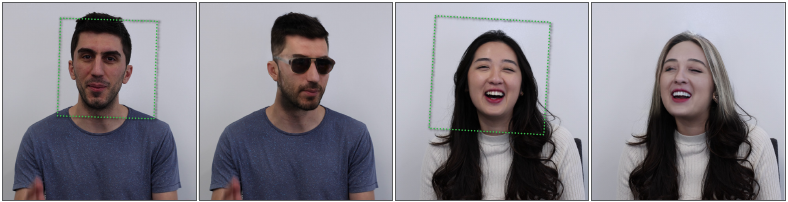} 
   \vspace{-6mm}
   \caption{\textbf{Spatial Consistency Additional Results.} \name composites images with challenging boundaries such as long hair \textit{(left)}, yielding faithful hair color change results. For hard boundary cases, such as matching with a static piece of hair outside the boundary crop \textit{(right)}, it plausibly connects the contents of the two images.}
   \label{fig:stitch-boundaries_sup}
   \vspace{-4mm}
\end{figure}

We showcase some supplemental qualitative results that demonstrate in further examples that VIVE3D is able to (1) apply existing latent space editing techniques such as InterfaceGAN~\cite{Shen2020InterFaceGAN} to generate natural-looking results (\Cref{fig:editing-directions_sup}) with performance comparable to previous 2D techniques, (2) create plausible image compositions for diverging from the original head angle, generalizing to various camera viewpoints given a source frame (\Cref{fig:view-directions_sup}), (3) generate high-quality results that are temporally consistent for combined angle and attribute editing (\Cref{fig:teaser_sup}), and (4) synthesize spatially consistent results (\Cref{fig:stitch-boundaries_sup}) even for challenging boundary cases.

\section{Optimization Details and Parameter Settings}
Our approach is implemented in Python 3.8 and uses PyTorch. We build our approach on the pretrained models and the publicly available codebase of EG3D~\cite{eg3d_code, Chan2021EG3D}. During our pipeline, we propose several optimization steps. Each of them is relying on ADAM as an optimizer. We run all experiments on a single NVIDIA A100 GPU and provide timings and hyperparameters for our various pipeline steps.
\begin{enumerate}[leftmargin=*]
\itemsep-0.1em 
\item{
    \textbf{Generator Inversion}: 
    For the initial inversion of the generator, we use a standard learning rate scheduler. We also ramp down the regularization weight of \offset{\idxF} from \inlinecode{\weight{wdist}} to \inlinecode{\weight{wdist\_target}}.\\
    \textit{Optimization hyperparameters}\\ 
    \inlinecode{\weight{L1} = 0.05},
    \inlinecode{\weight{face} = 1.0},
    \inlinecode{\weight{LPIPS} = 0.75},\\
    \inlinecode{\weight{wdist} = 0.05},
    \inlinecode{\weight{wdist\_target} = 0.005},\\
    \inlinecode{initial\_learning\_rate = 1e-2},
    \inlinecode{num\_steps = 600}\\
    \textit{Duration} 3~min 33~sec (5 target images)
}
\item{
    \textbf{Generator Fine-Tuning}: 
    We fine-tune the weights of the StyleGAN2 backbone of EG3D as well as the neural renderer, leaving learned weights of the Upsampling module untouched.\\
    \textit{Optimization hyperparameters}\\ 
    \inlinecode{\weight{L1} = 1.0}, 
    \inlinecode{\weight{LPIPS} = 0.3}, 
    \inlinecode{learning\_rate = 1e-3},\\ 
    \inlinecode{num\_steps = 300}\\
    \textit{Duration} 3~min 27~sec (5 target images)
}

\item{
    \textbf{Video Inversion}:
    During this optimization, we run a frame-per-frame inversion, starting from the average offset of all offsets \offset{\idxF} discovered in step 1. Using this strategy, we invert the first frame for \inlinecode{init\_num\_steps}. Each consecutive frame is started from the previous offset and optimized for \inlinecode{num\_steps}. We provide an early stopping criterion \inlinecode{loss\_threshold} and finish the current frame optimization in case the total loss falls below this threshold.\\
    \textit{Optimization hyperparameters}\\ 
    \inlinecode{\weight{L1} = 0.25},
    \inlinecode{\weight{face} = 1.2},
    \inlinecode{\weight{LPIPS} = 1.0},\\
    \inlinecode{\weight{wdist} = 0.01},
    \inlinecode{learning\_rate = 1e-2},\\
    \inlinecode{loss\_threshold = 0.25},\\
    \inlinecode{init\_num\_steps = 200}, 
    \inlinecode{num\_steps = 50}\\
    \textit{Duration} 19~sec (first frame), \mytilde4~sec/frame (consecutive frames)
}
\item{
    \textbf{Optical Flow Evaluation}:
    During this step, we evaluate the flow between the source face crop and the (angle-edited) target face to estimate the correction of the source crop needed to achieve a plausible inset composition.\\
    To estimate the flow, we use Farneb\"{a}ck optical flow with the following parameters:
    \inlinecode{pyr\_scale = 0.5},
    \inlinecode{levels = 8},
    \inlinecode{winsize = 25},\\
    \inlinecode{iterations = 7}, 
    \inlinecode{poly\_n = 5}, 
    \inlinecode{poly\_sigma = 1.2}\\
    \textit{Duration} \mytilde0.4~sec/frame
}
\item{
    \textbf{Inset Optimization}:
    In this optimization step, we can specify sizes \inlinecode{border\_size} for the width of the segmentation boundary region that is optimized and \inlinecode{edge\_size} to provide an offset distance for the border from the image boundary. We can again specify an early stopping criterion \inlinecode{border\_loss\_threshold} to stop when the border loss falls under this threshold, which in practice provides significant speedup.\\
    \textit{Optimization hyperparameters}\\ 
    \inlinecode{weight\_foreground = 1.0},
    \inlinecode{weight\_border = 2.0},\\
    \inlinecode{edge\_size = 50},
    \inlinecode{border\_size = 50},\\
    \inlinecode{num\_steps = 150},
    \inlinecode{learning\_rate = 1e-2},\\
    \inlinecode{border\_loss\_threshold = 0.05}\\
    \textit{Duration} \mytilde4~sec/frame (\mytilde16~sec/frame w/o early stopping)
}
\end{enumerate}

\section{Social Impact}
The ability to provide editability/customization in videos of humans has been an active area of research over the past few years. On one hand, it can have key applications in providing people the ability to express themselves in different ways (\eg, the ability to change hair color, add glasses, etc) during video calls, or more broadly in how they interact in the digital world. At the same time, such techniques introduce use cases for potentially malicious use that are worth discussing. For example, the ability to replace someone's face in a video from another person resembles deepfakes and could be used by bad actors. While the results such as what is shown in Figure~\ref{fig:head-change} are still not at the level that would be perceived as indistinguishable from an original video this is an important conversation to be had regardless. We encourage the interested reader to refer to concurrent work~\cite{dolhansky2020deepfake, guarnera2020deepfake, nguyen2022deep, agarwal2019protecting} for deep fake detection to discover edited videos.

\bibliographystyle{ieee_fullname}
\bibliography{bibliography}